\documentclass{article}

\usepackage[final]{corl_2026} 

\usepackage{amsmath}
\usepackage{amssymb}

\usepackage{subcaption}

\usepackage[inline]{enumitem}
\usepackage{multirow}

\usepackage{graphicx}
\usepackage{scalerel}
\usepackage{hyperref}
\usepackage{float}

\hypersetup{
    pdfauthor={Yilong Song, Harshvardhan Chandirasekar, Karthik Desingh, Chen Zhu-Tian},
    pdftitle={CrowdMAPS: Crowdsourcing Multiple Actions per State for Real-World Robot Imitation Learning}
}

\usepackage{booktabs}

\usepackage{wrapfig}

\title{CrowdMAPS\,\raisebox{-0.15em}{\includegraphics[height=1em, keepaspectratio]{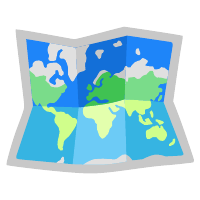}}: Crowdsourcing Multiple Actions per State for Real-World Robot Imitation Learning}
\author{
Yilong Song,\, Harshvardhan Chandirasekar,\, Karthik Desingh,\, and Chen Zhu-Tian \\
Department of Computer Science \& Engineering\\
  University of Minnesota, Twin Cities \\
  \texttt{\{song0837, chand863, kdesingh, ztchen\}@umn.edu} \\
}

\begin{document}
\maketitle


\begin{figure}[htbp]
    \centering
    \includegraphics[width=\textwidth]{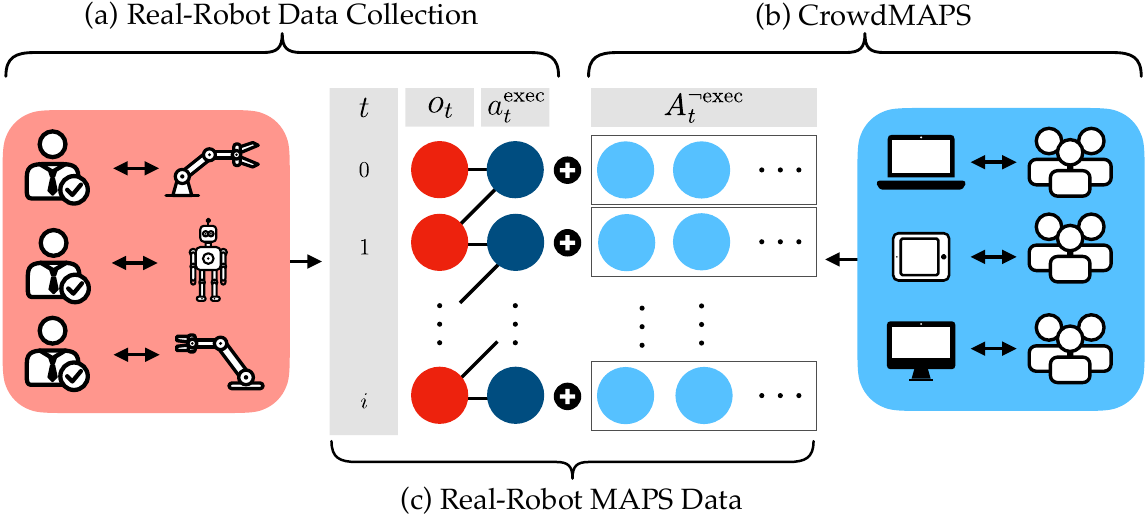}   
    \caption{
    \textbf{CrowdMAPS overview.}
    After standard real-robot data collection (a), CrowdMAPS (b) distributes visited states to remote annotators, who provide unexecuted alternative actions through a web interface. 
    The resulting real-robot Multiple Actions per State (MAPS) data (c) enables higher policy success rates and sample efficiency. Please visit project website \url{https://yilongsong.github.io/CrowdMAPS/} for more details.
    }
    \label{fig:CrowdMAPS}
\end{figure} 

\begin{abstract}
    Real-world robot imitation learning is often limited by the cost of collecting physical demonstrations, which typically yield only one action label per visited state. We present CrowdMAPS, a web-based crowdsourcing paradigm for augmenting real-robot datasets with Multiple Actions per State (MAPS). After or during standard robot data collection, CrowdMAPS distributes visited robot states to remote annotators through a web-based 3D interface, where they specify alternative valid actions without requiring access to physical robot hardware. This expands robot datasets along a complementary ``width'' dimension---by increasing the number of actions per visited state---rather than only by collecting more trajectories. We evaluate CrowdMAPS using keyframe-centric goal-pose annotations supplied by non-expert annotators and train imitation policies on the resulting MAPS data. Across five real-world manipulation tasks, increasing MAPS width from one to ten actions per state improves sample efficiency and increases average success rates by 35.6 percentage points for keypose policies and 33.0 percentage points for trajectory policies. These results show that software-mediated human annotations can augment physical robot data to afford denser supervision, establishing CrowdMAPS as a scalable digital layer for real-robot data collection.
\end{abstract}

\keywords{Real-World Robot Data Collection, Imitation Learning}

\section{Introduction}

Robot learning needs large, diverse, high-quality datasets, yet real-robot data remains expensive because each demonstration often requires physical execution, specialized hardware, and real-time human control~\citep{Padalkar2024, Brohan2023}.  In language and vision, large-scale datasets leverage internet-scale human activity and web-based annotation~\citep{Brown2020, Radford2021, vonAhn2008recaptcha, Deng2009, Lin2014}. Inspired by this success, recent systems crowdsource robot data by allowing remote users to teleoperate real robots through web interfaces~\citep{Mandlekar2018, Mirchandani2025RoboCrowd, Mirchandani2025RoboCade, Agarwal2026COBALT}.

However, robot crowdsourcing is inherently constrained by embodiment. One robot can execute only one action from one physical state at a time. Existing web-based systems therefore follow a synchronous ``one user controls one robot'' model, where every contribution requires physical hardware and throughput remains bounded by the number of robots and trajectory execution time. When contributors outnumber robots, users wait; moreover, noisy demonstrations can be less sample-efficient than expert data. Web-based teleoperation thus broadens who can contribute without removing the physical execution bottleneck.

We approach this bottleneck from a different angle: \textbf{instead of using the crowd only to collect more trajectories, can we use the crowd to extract more supervision from each trajectory already collected?} 
Our key observation is that a single visited robot state often admits multiple valid next actions, especially in manipulation tasks with geometric redundancy, multimodal task orderings, or alternative feasible grasps. Standard imitation-learning datasets record only the action that happened to be executed, leaving these other valid actions unobserved. This suggests a different decomposition of robot data collection: 
real robots are needed to acquire physically grounded states, but additional valid actions for those states can be annotated asynchronously in software.

We present \textbf{CrowdMAPS}, a web-based crowdsourcing paradigm for augmenting real-robot datasets with \textbf{Multiple Actions per State} (MAPS). 
During or after standard robot data collection, CrowdMAPS distributes visited robot states to remote non-expert annotators through a web-based 3D interface, where they specify alternative valid goal-pose actions without requiring access to physical robot hardware. 
This process performs \emph{action densification}: it expands robot datasets along a complementary ``width'' dimension by increasing the number of valid actions associated with each visited state, rather than only increasing the number of collected trajectories.

Across five real-world table-top manipulation tasks, increasing MAPS width from one to ten actions per keyframe improves sample efficiency with respect to robot time and increases average keypose-policy success rates by 35.6 percentage points, with larger gains on longer-horizon and more multimodal tasks. For trajectory policies, co-training with width-ten MAPS data improves average success by 33.0 percentage points and consistently outperforms adding an equal or greater number of expert teleoperated trajectories. These results show that action densification can improve imitation learning without proportionally increasing hardware-bound data collection.

Our contributions are:
\begin{enumerate}[leftmargin=*, topsep=-3pt, itemsep=0pt, parsep=0pt, partopsep=0pt]
    \item \textbf{CrowdMAPS}, an asynchronous crowdsourcing framework that decouples hardware-bound state acquisition from software-scalable action annotation, bypassing the one-user-one-robot bottleneck of synchronous web teleoperation.
    \item \textbf{Multiple Actions per State (MAPS)}, a data structure for imitation learning that densifies supervision by associating each visited state with multiple valid action labels.
    \item An empirical evaluation across five real-world manipulation tasks and two policy formulations, showing that increasing MAPS width improves policy success rates and sample efficiency, especially for longer-horizon and multimodal tasks, and that MAPS data can outperform adding an equal or greater number of expert teleoperated trajectories.
\end{enumerate}
\section{Related Work}
\label{sec:related_work}
\textbf{Real-Robot Data Collection:} Existing methods for real-robot data collection fall into three main categories: kinesthetic teaching~\citep{Argall2009}, teleoperation~\citep{Zhu2022VIOLA, Mandlekar2018, Brohan2023, Duan2023AR2D2, Wang2024EVE, Qin2023AnyTeleop, Wu2024GELLO, Zhao2023ALOHA}, and human demonstrations~\citep{Chi2024, Xu2025, Wang2024DexCap, Bahl2022, Lepert2025Phantom, lepert2025masquerade}. CrowdMAPS is compatible with and complementary to these paradigms: during or after physical data collection, CrowdMAPS distributes visited states to remote annotators to collect additional action labels per state, expanding training data offline and hardware free. 

\textbf{Crowdsourcing Real-Robot Data:} Crowdsourcing is a natural strategy for scaling labor intensive data, as it enables access to a large, parallelized workforce at a low marginal cost. Prior frameworks proposed systems, interfaces, incentives, and demonstrated the feasibility of crowdsourcing real-robot data, but maintained the standard ``one user controls one robot at one time'' scheme, capping the number of concurrent users by the number of available physical robots~\citep{Mandlekar2018, Muller2024OpenBotFleet, Mirchandani2025RoboCade, Agarwal2026COBALT}. In comparison, CrowdMAPS better leverages the core benefit of crowdsourcing---the scalability of labor---by allowing an arbitrary number of remote users to contribute action labels concurrently.

\textbf{Policy Learning with Multiple Actions per State:} Using many-actions or all-actions per state for policy learning has been thoroughly explored in Reinforcement Learning (RL). The Policy Gradient Theorem dictates that the true gradient contains an expectation over the entire action space~\citep{sutton2018reinforcement}; analytically via the law of total variance, MAPS can be shown to yield variance reduction in gradient estimation~\citep{kool2020estimating, nauman2023many}. Motivated by the theoretical guarantees, prior RL work has developed a lineage of all- or multi-action estimators~\citep{asadi2017mean, ciosek2018expected, petit2019all, kool2020estimating, nauman2023many}. While MAPS provides similar variance reduction benefits for imitation learning (App.~\ref{app:MAPS_variance_reduction_proof}), procuring alternative action labels has hitherto been logistically infeasible. To the best of our knowledge, CrowdMAPS is the first work to collect and leverage MAPS data for imitation learning.
\section{CrowdMAPS}
\label{sec:method}

CrowdMAPS augments real-robot imitation-learning datasets by collecting multiple valid action labels for states that have been visited by a physical robot. Let $\mathcal{O}$ and $\mathcal{A}$ denote the observation and action spaces. A standard executed dataset contains trajectories
\begin{equation}
\mathcal{D}_{\mathrm{exec}}
=
\{\tau_i^{\mathrm{exec}}\}_{i=1}^{N},
\quad \text{with} \quad
\tau_i^{\mathrm{exec}}
=
\left((o_t, a_t^{\mathrm{exec}})\right)_{t=0}^{T_i},
\end{equation}
where $a_t^{\mathrm{exec}}$ is the action that physically transitions the robot from $o_t$ to $o_{t+1}$. CrowdMAPS converts this dataset into a Multiple-Actions-per-State (MAPS) dataset:
\begin{equation}
\mathcal{D}_{\mathrm{MAPS}}
=
\{\tau_i^{\mathrm{MAPS}}\}_{i=1}^{N},
\quad \text{with} \quad
\tau_i^{\mathrm{MAPS}}
=
\left((o_t, A_t)\right)_{t=0}^{T_i},
\quad
A_t = \{a_t^{\mathrm{exec}}\} \cup A_t^{\neg\mathrm{exec}},
\end{equation}

where $A_t^{\neg\mathrm{exec}}$ is the set of valid but unexecuted alternative action labels. We call $K_t = |A_t|$ the action density of state $o_t$. When a target density $K$ is maintained over a selected subset of states, we refer to $K$ as the dataset ``width.'' Intuitively, MAPS data provides a denser empirical sampling of the valid action distribution $\pi^*(a \mid o_t)$ than standard single-action demonstrations, while preserving the same set of physically visited observations (see qualitative results and analysis in App.~\ref{app:gain_analysis}). Analytically, under independent sampling assumptions, MAPS reduces the action-dependent variance of the gradient estimator (see proof in App.~\ref{app:MAPS_variance_reduction_proof}). 
\subsection{Keyframe-Centric Goal-Pose Instantiation}
\label{subsec:action_space}

The generic MAPS formulation allows different action representations. In our implementation, each action label is represented as a \textbf{\emph{goal-pose action}}:
$
g = (v, q, c) \in SE(3) \times \{0,1\},
$
where $v \in \mathbb{R}^3$ is the target end-effector translation, $q \in SO(3)$ is the target orientation, and $c \in \{0,1\}$ is a binary gripper command. These goal-pose actions are fulfilled by a motion planner. Thus, $g$ specifies the next key end-effector state, such as a grasp pose, rather than a dense sequence of waypoints or lower-level control commands.

\begin{figure}[th]
    \centering
    \setlength{\fboxsep}{5pt} 
    \setlength{\fboxrule}{0.5pt}

    \begin{subfigure}[b]{0.40\linewidth}
        \centering
        \caption{Standard Trajectory Dataset}
        \includegraphics[height=4cm]{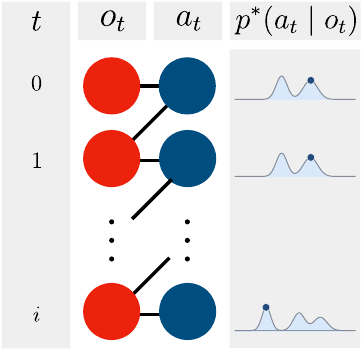}
        \label{fig:std}
    \end{subfigure}
    \hfill 
    \begin{subfigure}[b]{0.55\linewidth}
        \centering
        \caption{MAPS Dataset (Keyframe)}
        \includegraphics[height=4cm]{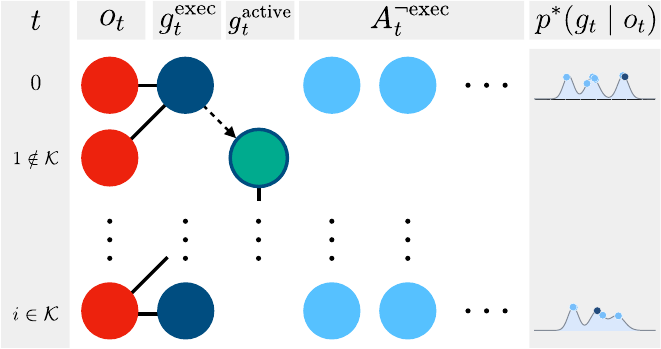}
        \label{fig:maps}
    \end{subfigure}

\caption{
\textbf{Standard trajectory data vs. keyframe-centric MAPS data.}
A standard real-robot trajectory records one executed action \(a_t\) for each visited observation \(o_t\). 
CrowdMAPS preserves the executed trajectory but augments selected keyframes \(t \in \mathcal{K}\) with additional valid, unexecuted goal-pose labels \(A_t^{\neg\mathrm{exec}}\). 
Thus, keyframes store \(A_t=\{g_t^{\mathrm{exec}}\}\cup A_t^{\neg\mathrm{exec}}\), while intermediate frames \(t\notin\mathcal{K}\) are paired only with the active executed goal (Sec.~\ref{subsec:action_space}). 
This increases action-label density per visited state without creating additional real-world transition branches.
}   
    \label{fig:MAPS}
\end{figure}

Our data scheme is keyframe-centric. \textbf{\emph{Keyframes}} ($t \in \mathcal{K}$) occur at trajectory start and upon goal completions; they store the executed goal pose and crowdsourced alternatives. \textbf{\emph{Intermediate frames}} occur while pursuing a goal and are paired only with the active executed goal:
\begin{equation}
A_t =
\begin{cases}
\{g_t^{\mathrm{exec}}\} \cup A_t^{\neg\mathrm{exec}}
& \text{if } t \in \mathcal{K}, \\
\{g_t^{\mathrm{active}}\}
& \text{if } t \notin \mathcal{K},
\end{cases}
\end{equation}
where $g_t^{\mathrm{exec}}$ is the executed goal-pose action selected at keyframe $t$, $A_t^{\neg\mathrm{exec}}$ is the set of valid unexecuted alternatives, and $g_t^{\mathrm{active}} = g_{\kappa(t)}^{\mathrm{exec}}$ denotes the most recent executed goal still being pursued at intermediate frame $t$. See Fig.~\ref{fig:MAPS} for a visualization of this data scheme. 
This representation makes MAPS practical to collect: keyframes concentrate annotation at task-level decision points, goal poses are easily specified through a web interface, and intermediate frames preserve closed-loop supervision using the active executed goal.

\subsection{Web-Based Crowdsourcing Procedure}
\label{subsec:framework}

\begin{figure}[t]
    \centering
    \includegraphics[width=\textwidth]{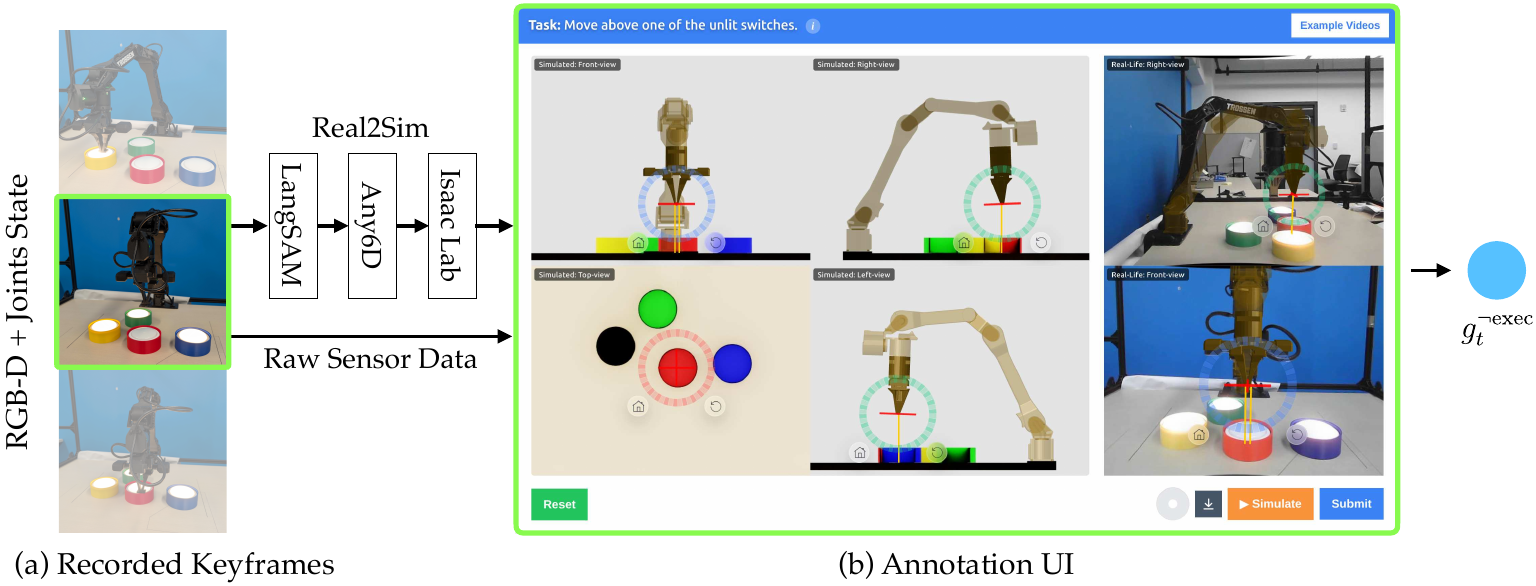}
    \caption{
\textbf{CrowdMAPS web annotation pipeline.}
Recorded real-robot keyframes (a) contain RGB-D and proprioceptive data. Using task-specific CAD models and object text descriptions, LangSAM segments the relevant objects, Any6D estimates their 6-DoF poses, and the corresponding meshes are placed in Isaac Lab to reconstruct the scene. The reconstructed scene is displayed alongside raw sensor views in the annotation interface (b), where annotators manipulate the rendered robot to specify an alternative goal-pose action, submitted as an unexecuted label \(g \in A_t^{\neg\mathrm{exec}}\).
}
    \label{fig:system}
\end{figure}

CrowdMAPS collects unexecuted alternative goal poses via a two-stage procedure:
\begin{enumerate}[leftmargin=*, topsep=0pt, itemsep=0pt]
    \item \textbf{State Acquisition.} An expert collects standard real-robot trajectories using the same goal-pose representation.
    \item \textbf{Action Densification.} Recorded keyframes are distributed to remote annotators. Using a 3D scene reconstruction and raw sensor views, annotators manipulate a rendered robot model to specify candidate alternative goal poses for each visited state (Fig.~\ref{fig:system}). Accepted alternatives $A_t^{\neg\mathrm{exec}}$ are combined with the executed action $g_t^{\mathrm{exec}}$ to form the final MAPS action set $A_t$.
\end{enumerate}

Each submission is manually validated by a researcher for consistency with the manipulation strategy established during physical data collection, semantic alignment with the keyframe instruction, and physical coherence. This process filters invalid or off-strategy actions while preserving valid diversity within the selected strategy. Further details are provided in App.~\ref{app:crowdsourcing_ui}.

CrowdMAPS enables simultaneous, hardware-free annotation by multiple users, avoiding the physical bottleneck of synchronous teleoperation. We formalize the broader crowdsourcing paradigm underlying this implementation in App.~\ref{app:general_crowdsourcing_formulation}.

\subsection{Using MAPS Data for Imitation Learning} 
\label{subsec:using_MAPS}

We showcase the utility of our keyframe-centric MAPS data under two policy-training settings.

\textbf{Learning keypose policies.}
Our keyframe-centric MAPS dataset naturally supports learning a keypose policy
$\pi_{\mathrm{keypose}}(g \mid O_t)$, where \(g\) is a goal-pose action and \(O_t\) denotes the policy conditioning context, including the current observation, proprioception, and recent history. Since each visited state \(o_t\) is associated with an action-label set \(A_t\), we supervise the policy over all labels in the set:
\begin{equation}
\label{eq:maps_objective}
\mathcal{J}_{\mathrm{keypose}}(\theta)
=
\mathbb{E}_{\tau \sim \mathcal{D}_{\mathrm{MAPS}}}
\left[
\frac{1}{Z_\tau}
\sum_{t=0}^{T}
\sum_{g \in A_t}
w_{t,g}
\mathcal{L}_{\mathrm{gen}}(\pi_\theta, g, O_t)
\right],
\end{equation}
where \(\mathcal{L}_{\mathrm{gen}}\) is the generative modeling loss, \(w_{t,g}\) is an action-type weight, and $Z_\tau = \sum_{t=0}^{T} \sum_{g \in A_t} w_{t,g}$ normalizes the weighted loss within a trajectory.

The weights distinguish three types of supervision: 
\emph{executed keyframe actions} \((t \in \mathcal{K}, g=g_t^{\mathrm{exec}})\), 
\emph{unexecuted keyframe alternatives} \((t \in \mathcal{K}, g \in A_t^{\neg\mathrm{exec}})\), 
and \emph{intermediate-frame active goals} \((t \notin \mathcal{K})\). 
We denote their weights as \(w_{\mathrm{key,exec}}\), \(w_{\mathrm{key,\neg exec}}\), and \(w_{\mathrm{inter}}\), respectively. These weights balance different supervision sources and help mitigate the MAPS shift issue discussed at the end of this section.
Implementation details are provided in App.~\ref{app:keypose_policy_details}.

\textbf{Co-training trajectory policies.}
MAPS data can also be used with standard trajectory policies, which predict time-aligned action chunks:
\begin{equation}
\label{eq:traj_policy}
\pi_{\mathrm{traj}}(\mathbf{a}_t \mid O_t, d),
\quad
\mathbf{a}_t = (a_t, a_{t+1}, \dots, a_{t+H-1}) \in \mathcal{A}^H,
\end{equation}
where \(d\) is an optional dataset-source indicator. Because MAPS provides single goal poses rather than dense action chunks, we convert each ($g \in A_t$) into a synthetic chunk ($\mathbf{a}^{\mathrm{MAPS}}t(g)$) by interpolating from the current end-effector pose to ($g$) over horizon ($H$), using the average velocity ($\bar{v}{\mathrm{batch}}$) of teleoperated trajectories in the training batch. Remaining steps after reaching the target hold the target pose.

We then co-train the trajectory policy on standard trajectory data \(\mathcal{D}_{\mathrm{traj}}\) and converted MAPS data by optimizing the expected loss over both datasets simultaneously:
\begin{equation}
\label{eq:traj_maps_objective}
\begin{aligned}
\mathcal{J}_{\mathrm{traj}}(\theta) =\, & \mathbb{E}_{\tau \sim \mathcal{D}_{\mathrm{traj}}} \left[ \sum_{t=0}^{T} w_{\mathrm{traj}} \mathcal{L}_{\mathrm{gen}}(\pi_\theta, \mathbf{a}_t, O_t) \right] \\
  + &\mathbb{E}_{\tau \sim \mathcal{D}_{\mathrm{MAPS}}} \left[ \frac{1}{Z_\tau} \sum_{t=0}^{T} \sum_{g \in A_t} w_{t,g} \mathcal{L}_{\mathrm{gen}}\big(\pi_\theta, \mathbf{a}^{\mathrm{MAPS}}_t(g), O_t\big) \right].
\end{aligned}
\end{equation}
By normalizing the combined weights (\(w_{\mathrm{traj}}\) and \(w_{t,g}\)) at the batch level during training, the policy learns a single coherent action distribution from both dense continuous demonstrations and sparse, multi-modal crowd labels. See App.~\ref{app:trajectory_policy_details} for implementation details.

\textbf{MAPS shift and action-type weighting.}
Using unexecuted alternatives introduces an important distributional issue. MAPS expands the conditional action-label distribution at visited states, but it does not expand the observation distribution with successor states induced by those alternatives:
\begin{equation}
p_{\mathcal{D}_{\mathrm{MAPS}}}(a \mid o) \neq p_{\mathrm{exec}}(a \mid o), \quad p_{\mathcal{D}_{\mathrm{MAPS}}}(o) = p_{\mathrm{exec}}(o).
\end{equation}
Thus, if a policy frequently selects unexecuted alternatives during rollout, it may visit states whose consequences are underrepresented in the training data. We refer to this issue as \emph{MAPS shift}.

We mitigate MAPS shift in two ways. First, the validation procedure in Sec.~\ref{subsec:framework} restricts crowdsourced actions to the researcher-selected manipulation strategy, reducing the risk of MAPS shift from off-strategy alternatives.
Second, we use action-type weighting
($w_{\mathrm{key,exec}} > w_{\mathrm{key,\neg exec}} > w_{\mathrm{inter}}$):
executed actions receive the highest weight because they correspond to observed transitions, unexecuted alternatives receive lower weight because they are not physically grounded, and intermediate goals receive the lowest weight because they represent in-progress motion rather than task-level decisions.
\section{Experiments}

We evaluate the central hypothesis that increasing action density per visited state improves imitation learning without requiring additional physical rollouts. We test this hypothesis in two settings: first in the native MAPS setting, where goal-pose labels directly supervise keypose policies; and then in a more standard trajectory-policy setting, where MAPS labels are converted into synthetic action chunks and co-trained with continuous expert demonstrations. We ask:

\begin{enumerate}[label=\textbf{RQ\arabic*.}, leftmargin=*, topsep=-3pt, itemsep=0pt, parsep=0pt, partopsep=0pt]
\item \textbf{Native MAPS supervision:} Does increasing MAPS width improve keypose policy success and sample efficiency?
\item \textbf{Transfer to trajectory policies:} Can high-width MAPS data improve trajectory policies and outperform additional expert teleoperated trajectories?
\end{enumerate}

\subsection{Setup}
\label{subsec:experiments_setup}

\textbf{Data collection.}
We recruit 47 participants to provide alternative goal-pose annotations using the procedure in Sec.~\ref{subsec:framework}. Most participants (78.7\%) have no prior experience with robot arms. Across the four tasks with timing logs, median submission times range from 7 to 12 seconds, while submission acceptance rates range from 51\% to 98\% across all five tasks. Further details on the interface, annotation protocol, and validation process are provided in App.~\ref{app:crowdsourcing_ui} and App.~\ref{app:crowdsourcing_process}.

\textbf{Tasks.}
We evaluate on five real-world manipulation task families: \textbf{\emph{Drawer}}, \textbf{\emph{Insertion}}, \textbf{\emph{Pouring}}, \textbf{\emph{Sorting}}, and \textbf{\emph{Switches}}. 
The tasks vary in horizon, precision, and multimodality. 
Drawer combines articulated-object manipulation and pick-and-place; Insertion requires precise alignment; Pouring requires a specialized transport-and-pour motion; Sorting is multimodal because the two cubes can be picked in different orders; and Switches is multimodal because the lights can be pressed in different orders. 
We use two Switches variants: Switches$^+$, where all four lights must be activated, and Switches$^-$, where three lights are already on and the policy presses the remaining light. Full task specifications are provided in App.~\ref{app:experiments}.

\textbf{Policy settings.}
We evaluate two policy formulations from Sec.~\ref{subsec:using_MAPS}. 
For \textbf{\emph{keypose policies}}, we train directly on MAPS goal-pose labels and vary the action density \(K\) per keyframe to answer RQ1. 
For \textbf{\emph{trajectory policies}}, we co-train diffusion policies on continuous expert demonstrations and converted MAPS action chunks, comparing MAPS data with additional expert teleoperated trajectories to answer RQ2. 
Additional implementation details and ablations are provided in App.~\ref{app:experiments}.

\textbf{Evaluation protocol.}
Each trained model is evaluated with 25 real-world rollouts under randomized initializations. Success is binary for all tasks except Switches$^+$, where we record the number of successfully activated lights across 25 rollouts, for a maximum score of 100.

\subsection{Findings} 
\label{subsec:experiment_results}
\begin{figure}[H]
    \centering

    \begin{minipage}{\textwidth}
    \centering
    \small
    \setlength{\tabcolsep}{5pt}
    \begin{tabular}{l c ccccc cccc cccc cccc}
        \toprule
        $K$ & Drawer & \multicolumn{5}{c}{Insertion} & \multicolumn{4}{c}{Pouring} & \multicolumn{4}{c}{Sorting} & \multicolumn{4}{c}{Switches$^-$} \\
        \cmidrule(lr){2-2} \cmidrule(lr){3-7} \cmidrule(lr){8-11} \cmidrule(lr){12-15} \cmidrule(lr){16-19}
        & 5 & 20 & 30 & 40 & 50 & 60 & 15 & 20 & 30 & 45 & 20 & 30 & 40 & 50 & 50 & 60 & 70 & 80 \\
        \midrule
        1 & 12 & 8 & 36 & 68 & 72 & 76 & 28 & 32 & 56 & 92 & 8 & 20 & 28 & 36 & 24 & 24 & 36 & 40 \\
        2 & 52 & \textbf{16} & 64 & 76 & 88 & 92 & 32 & 40 & 72 & 96 & 4 & 8 & 48 & 52 & 28 & 44 & 52 & 60 \\
        5 & \textbf{76} & 8 & 64 & 76 & 88 & 92 & 32 & 40 & 80 & 96 & 20 & 28 & 64 & 72 & 36 & 40 & 60 & 80 \\
        10 & \textbf{76} & 12 & \textbf{72} & \textbf{92} & \textbf{96} & \textbf{96} & \textbf{56} & \textbf{84} & \textbf{96} & \textbf{100} & \textbf{28} & \textbf{40} & \textbf{88} & \textbf{92} & \textbf{56} & \textbf{80} & \textbf{84} & \textbf{88} \\
        \midrule
        10 (noise) & 4 & 12 & 68 & 88 & 68 & 52 & 32 & 16 & 52 & 80 & 0 & 8 & 60 & 24 & 40 & 56 & 52 & 68 \\
        \bottomrule
    \end{tabular}
    \end{minipage}

    \vspace{0.5em}

    \hspace*{0.13\textwidth}
    \includegraphics[width=0.83\textwidth]{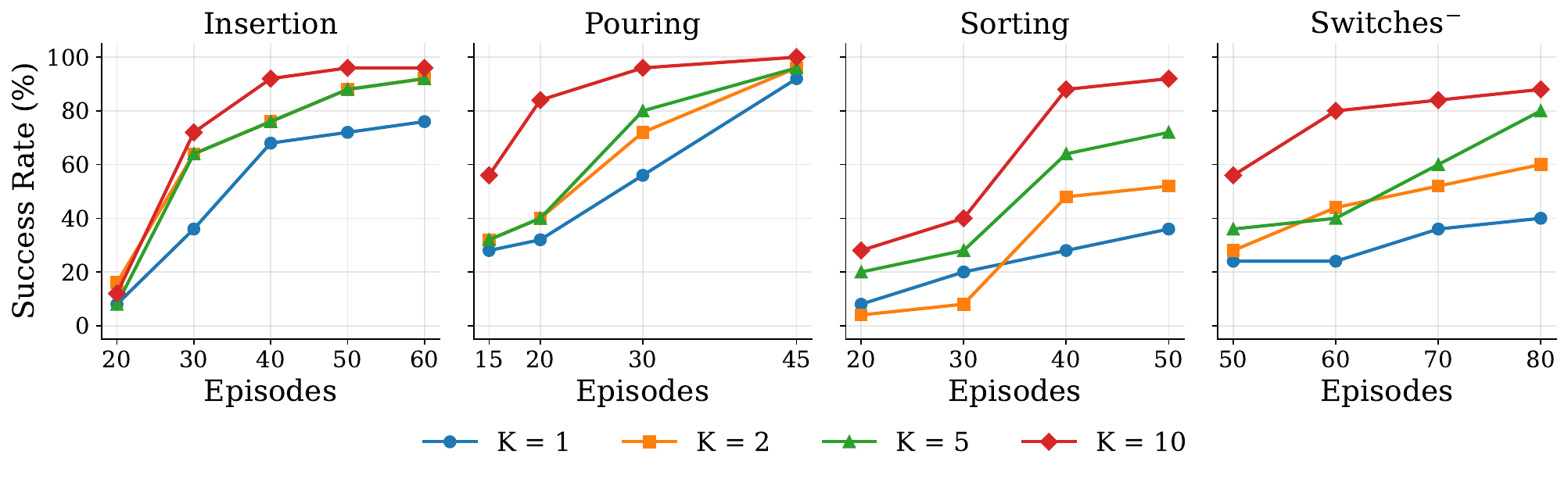}

    \caption{
    \textbf{Keypose policy performance as MAPS width increases.}
    Top: success rates (\%) for keypose policies trained with different MAPS widths \(K\), where columns under each task indicate the number of training trajectories.
    The \(K=10\) noise baseline replaces crowdsourced alternatives with Gaussian perturbations of executed actions, with covariance estimated from crowd annotations (App.~D.4).
    Bottom: scaling trends for crowdsourced MAPS widths on Insertion, Pouring, Sorting, and Switches$^-$.
    Each entry is evaluated over 25 real-world rollouts with randomized initializations; best results are in bold.
    }    
    \label{fig:keypose_results}
\end{figure}

\textbf{RQ1: Wider MAPS data improves keypose policies, especially on longer-horizon and multimodal tasks.}
Fig.~\ref{fig:keypose_results} shows that increasing MAPS width improves both success rate and sample efficiency. Averaged across all reported task-episode settings, increasing width from \(K=1\) to \(K=10\) improves success from \(38.7\%\) to \(74.2\%\), a gain of \(35.6\) percentage points. The effect is strongest on tasks with longer horizons or greater multimodality. For Sorting and Switches$^-$, increasing \(K\) from 1 to 10 yields an average gain of \(42.5\) percentage points, and \(K=10\) is the only setting that reaches above \(85\%\) success within the evaluated episode ranges.

\begin{figure}[H]
    \centering

    \begin{minipage}[c]{0.09\linewidth}\strut\end{minipage}%
    \hfill
    \begin{minipage}[c]{0.89\linewidth}
        \centering\small
        \makebox[0.25\linewidth][c]{$K=1$}%
        \makebox[0.25\linewidth][c]{$K=2$}%
        \makebox[0.25\linewidth][c]{$K=5$}%
        \makebox[0.25\linewidth][c]{$K=10$}%
    \end{minipage}

    \par\vspace{2pt}

    \begin{minipage}[c]{0.09\linewidth}
        \centering\footnotesize (a)\\Insertion
    \end{minipage}%
    \hfill
    \begin{minipage}[c]{0.89\linewidth}
        \includegraphics[width=\linewidth]{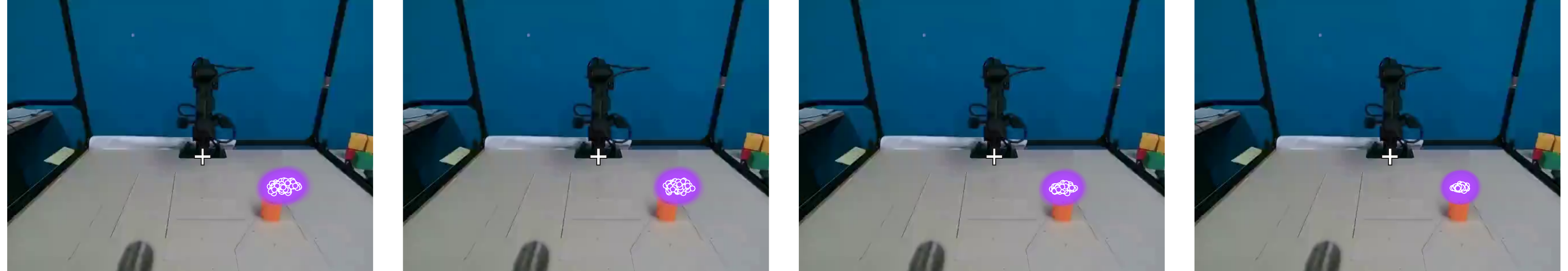}
    \end{minipage}

    \par\vspace{4pt}

    \begin{minipage}[c]{0.09\linewidth}
        \centering\footnotesize (b)\\Sorting
    \end{minipage}%
    \hfill
    \begin{minipage}[c]{0.89\linewidth}
        \includegraphics[width=\linewidth]{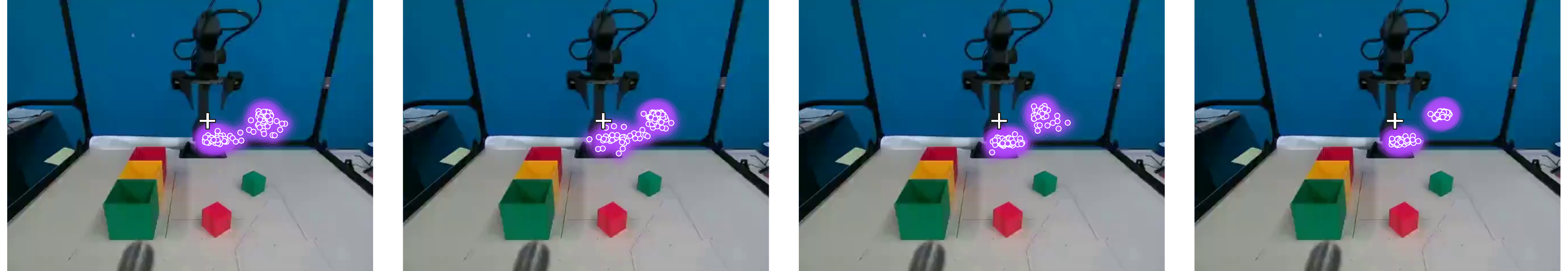}
    \end{minipage}

    \caption{
    \textbf{Policy predictions at held-out keyframes.}
    We sample 100 goal-pose predictions from keypose policies trained with MAPS widths $K=1,2,5,10$ at the same held-out observation.
    Purple markers denote predictions; the white cross denotes the current gripper position.
    (a) Insertion: increasing $K$ reduces both the offset and dispersion of the predictions.
    (b) Sorting: increasing $K$ results in sharper, more clearly separated valid action modes.
    }
    \label{fig:policy_preds}
\end{figure}

The gains are smaller on shorter or easier tasks once all variants begin to converge. For example, Insertion and Pouring show less separation at larger data sizes, indicating that additional action density is most valuable when the task involves substantial action variety or longer-horizon decision making. Overall, these results support the central hypothesis: increasing the number of valid action labels per visited state can improve policy learning without collecting more physical trajectories.

We additionally compare against a strong Gaussian-noise baseline, calibrated using the within-mode variance of crowdsourced actions (App.~\ref{app:noise_baseline}). Despite matching $K=10$ in action density, the noise baseline achieves only $43.3\%$ average success, compared to $74.2\%$ with crowdsourced labels, and exhibits negative scaling in several settings (more demos, worse performance). These results suggest that the gains cannot be replicated simply by adding Gaussian-perturbed action labels.

To qualitatively assess the benefits of wider MAPS data, we visualize policy predictions at held-out keyframes (Fig.~\ref{fig:policy_preds}). For the unimodal Insertion task, increasing $K$ produces progressively sharper predictions around the valid goal pose, illustrating the benefit of within-mode densification. For the multimodal Sorting task, increasing $K$ produces both sharper individual modes and better separation between distinct modes. Together, these observations suggest that increasing MAPS width improves learned action distributions through both denser within-mode sampling and multimodal coverage. Further analysis is provided in App.~\ref{app:gain_analysis}.

\begin{table*}[th]
    \centering
    \small

\caption{
\textbf{Trajectory policy performance under different data augmentation strategies.}
All methods start from \(E_{\text{base}}\). 
We compare adding expert trajectories \(E_{\text{added}}\) with adding MAPS data \(\mathcal{D}_{\text{MAPS}}\) at different widths \(K\). 
Results are success rates over 25 rollouts, except Switches$^+$, reported as activated lights out of 100. 
Best results are in bold.
}
    \label{tab:trajectory_results}
\begin{tabular}{l ccccc}
\toprule
\textbf{Training Data} & \textbf{Drawer} & \textbf{Insertion} & \textbf{Pouring} & \textbf{Sorting} & \textbf{Switches$^+$} \\
\midrule
Data sizes: $E_{\text{base}} / E_{\text{added}} / \mathcal{D}_{\text{MAPS}}$ 
& 10 / 5 / 5 & 200 / 100 / 60 & 50 / 50 / 45 & 50 / 50 / 50 & 80 / 80 / 80 \\
\midrule
$E_{\text{base}}$ & 36 & 0 & 48 & 16 & 29 \\
$E_{\text{base}} + E_{\text{added}}$ & 56 & 24 & 80 & 52 & 54 \\
\midrule
$E_{\text{base}} + \mathcal{D}_{\text{MAPS}}(K=1)$ & 36 & 16 &76 & 32 & 41 \\
$E_{\text{base}} + \mathcal{D}_{\text{MAPS}}(K=2)$ & 48 & 28 & 84 & 60 & 53 \\
$E_{\text{base}} + \mathcal{D}_{\text{MAPS}}(K=5)$ & 52 & 24 & \textbf{92} & 80 & 60 \\
$E_{\text{base}} + \mathcal{D}_{\text{MAPS}}(K=10)$ & \textbf{76} & \textbf{40} & \textbf{92} & \textbf{88} & \textbf{70} \\
\bottomrule
\end{tabular}
\end{table*}

\textbf{RQ2: The action-densification benefit transfers to trajectory policies and can exceed additional expert teleoperation.}
For trajectory policies, we compare two ways of augmenting a base expert dataset \(E_{\text{base}}\): collecting additional expert teleoperated trajectories \(E_{\text{added}}\), or adding CrowdMAPS data \(\mathcal{D}_{\text{MAPS}}\) with increasing action density \(K\). This comparison tests whether lower-fidelity, keyframe-centric MAPS supervision can compensate for not being dense continuous teleoperation by increasing the number of valid action labels per visited state.

Table~\ref{tab:trajectory_results} shows two patterns. 
First, adding MAPS data improves trajectory-policy learning over \(E_{\text{base}}\) across tasks, indicating that keyframe-centric MAPS labels provide useful supervision even after being converted into synthetic action chunks. The magnitude of the gain varies across tasks, likely reflecting both task difficulty and the relative amount of MAPS data added. 
Second, and more importantly, increasing \(K\) makes MAPS data substantially more effective. At \(K=1\), MAPS data often underperforms additional continuous expert trajectories, suggesting that single-label keyframe supervision has lower temporal fidelity than expert teleoperation. However, as \(K\) increases, the denser action supervision compensates for this limitation. At \(K=10\), \(E_{\text{base}} + \mathcal{D}_{\text{MAPS}}\) achieves the best performance on all evaluated tasks and outperforms \(E_{\text{base}} + E_{\text{added}}\). We note that this comparison approximately matches aggregate robot occupancy (2.2 vs.\ 2.3 hours for \(\mathcal{D}_{\text{MAPS}}\) and \(E_{\text{added}}\), respectively), but not total human effort: at matched performance, \(\mathcal{D}_{\text{MAPS}}\) requires approximately \(1.5\times\) greater total human time. Thus, Table~\ref{tab:trajectory_results} demonstrates augmentation effectiveness under comparable robot-time budgets, rather than general cost efficiency. These results suggest that MAPS is not merely additional data; its benefit comes from densifying the valid action distribution at already visited states.

\textbf{Takeaway.}
Across both policy formulations, the results show that MAPS improves imitation learning. The keypose experiments demonstrate the direct benefit of increasing MAPS width, while the trajectory-policy experiments show that this benefit transfers beyond the native goal-pose formulation and can outperform adding more expert teleoperated trajectories. These findings position action densification as a complementary scaling axis for real-robot data collection.

\section{Limitations and Future Work}

\textbf{Action representation and task scope.}
CrowdMAPS uses goal-pose actions, which enable asynchronous web annotation but are less expressive than dense teleoperation and require motion planning. Our evaluation includes articulated-object manipulation and Pouring but focuses on quasi-static, single-arm tabletop tasks. Dynamic, deformable, strongly contact-rich, and bimanual tasks may require interfaces for short motion segments, contact constraints, or hierarchical commands.

\textbf{Real-to-sim reconstruction.}
The current pipeline requires task-specific CAD models and object descriptions provided manually, and thus incurs setup costs for new tasks and objects. More generalizable reconstruction methods could reduce this dependence and extend the annotation interface to less structured environments.

\textbf{Training design choices.}
Holding total keyframe weight constant as $K$ increases shifts more influence toward crowdsourced alternatives; other weighting schemes remain unablated. For trajectory policies, converting individual goal poses into dense action chunks relies on heuristic interpolation and may partly explain weaker performance at low $K$. The relative contributions of weighting and conversion warrant systematic investigation.

\textbf{MAPS shift and unexecuted actions.}
Because alternative actions are not physically executed, their successor observations are absent from training data, potentially inducing MAPS shift during rollout. Manual validation and action weighting mitigate this risk but restrict annotation to researcher-selected strategies. Future work could explore DAgger-style iterative crowdsourcing, model-based successor prediction, and synthetic-state annotation. Whether broader state coverage permits safely relaxing these constraints remains an open question.

\textbf{Reducing human validation effort.}
Our implementation relies on researchers to assess task relevance and physical feasibility. Experienced contributors, aggregated annotator judgments, or learned models could filter invalid actions and prioritize uncertain submissions for expert review. This could reduce administrative effort without treating unexecuted actions as automatically valid.

\textbf{Alternative MAPS sources.}
Crowdsourcing is only one mechanism for action densification. Alternatives include interfaces that collect several nearby goal poses at once, aggregating distinct executed actions at similar states in offline robot datasets, and synthesizing candidate actions using task symmetries or vision-language models. Such candidates would still require feasibility checks.

\section{Conclusion}

We presented \textbf{CrowdMAPS}, a web-based crowdsourcing paradigm for scaling real-robot imitation-learning data through \emph{action densification}. CrowdMAPS decouples hardware-bound state acquisition from software-scalable action annotation: real robots provide visited states, while remote annotators provide multiple valid goal-pose actions for those states. Across real-world manipulation tasks and two policy formulations, increasing MAPS width improves success rates and sample efficiency; in the trajectory-policy setting, high-width MAPS data can outperform adding more expert teleoperated trajectories. These results establish action densification as a practical complementary pathway for scaling real-world robot imitation learning.




\clearpage
\acknowledgments{We thank the anonymous reviewers and the Area Chair for their thoughtful feedback and constructive suggestions, which helped improve the clarity and completeness of this work.}


\bibliography{references}  

@InProceedings{Brown2020,
  title     = {Language Models are Few-Shot Learners},
  author    = {Tom B. Brown and Benjamin Mann and Nick Ryder and Melanie Subbiah and Jared Kaplan and Prafulla Dhariwal and Arvind Neelakantan and Pranav Shyam and Girish Sastry and Amanda Askell and Sandhini Agarwal and Ariel Herbert-Voss and Gretchen Krueger and Tom Henighan and Rewon Child and Aditya Ramesh and Daniel M. Ziegler and Jeffrey Wu and Clemens Winter and Christopher Hesse and Mark Chen and Eric Sigler and Mateusz Litwin and Scott Gray and Benjamin Chess and Jack Clark and Christopher Berner and Sam McCandlish and Alec Radford and Ilya Sutskever and Dario Amodei},
  booktitle = {Advances in Neural Information Processing Systems (NeurIPS)},
  year      = {2020},
  volume    = {33},
  pages     = {1877--1901}
}

@InProceedings{Radford2021,
  title     = {Learning Transferable Visual Models From Natural Language Supervision},
  author    = {Alec Radford and Jong Wook Kim and Chris Hallacy and Aditya Ramesh and Gabriel Goh and Sandhini Agarwal and Girish Sastry and Amanda Askell and Pamela Mishkin and Jack Clark and Gretchen Krueger and Ilya Sutskever},
  booktitle = {Proceedings of the 38th International Conference on Machine Learning (ICML)},
  year      = {2021},
  pages     = {8748--8763},
  volume    = {139}
}

@InProceedings{Padalkar2024,
  title     = {Open {X-E}mbodiment: Robotic Learning Datasets and {RT-X} Models},
  author    = {Abhishek Padalkar and Acorn Pooley and Ajinkya Jain and Alex Bewley and Alex Herzog and Alex Irpan and Alexander Khazatsky and Arthur Simon and Austin Ray and Brijen Thananjeyan and Chelsea Finn and others},
  booktitle = {Proceedings of the IEEE International Conference on Robotics and Automation (ICRA)},
  year      = {2024}
}

@InProceedings{Brohan2023,
  title     = {{RT-1}: Robotics Transformer for Real-World Control at Scale},
  author    = {Anthony Brohan and Noah Brown and Justice Carbajal and Yevgen Chebotar and Joseph Dabis and Chelsea Finn and Keerthana Gopalakrishnan and Karol Hausman and Alex Herzog and Jasmine Hsu and Julian Ibarz and Brian Ichter and Alex Irpan and Tomas Jackson and others},
  booktitle = {Robotics: Science and Systems (RSS)},
  year      = {2023}
}

@InProceedings{Chi2024,
  title     = {Universal Manipulation Interface: In-The-Wild Robot Teaching Without In-The-Wild Robots},
  author    = {Cheng Chi and Zhenjia Xu and Chuer Pan and Eric Cousineau and Benjamin Burchfiel and Siyuan Feng and Russ Tedrake and Shuran Song},
  booktitle = {Robotics: Science and Systems (RSS)},
  year      = {2024}
}

@InProceedings{Xu2025,
  title     = {{DexUMI}: Using Human Hand as the Universal Manipulation Interface for Dexterous Manipulation},
  author    = {Mengda Xu and Han Zhang and Yifan Hou and Zhenjia Xu and Linxi Fan and Manuela Veloso and Shuran Song},
  booktitle = {Conference on Robot Learning (CoRL)},
  year      = {2025}
}

@InProceedings{Deng2009,
  title     = {ImageNet: A Large-Scale Hierarchical Image Database},
  author    = {Jia Deng and Wei Dong and Richard Socher and Li-Jia Li and Kai Li and Li Fei-Fei},
  booktitle = {Proceedings of the IEEE Conference on Computer Vision and Pattern Recognition (CVPR)},
  year      = {2009},
  pages     = {248--255}
}

@InProceedings{Lin2014,
  title     = {Microsoft {COCO}: Common Objects in Context},
  author    = {Tsung-Yi Lin and Michael Maire and Serge Belongie and James Hays and Pietro Perona and Deva Ramanan and Piotr Doll{\'a}r and C. Lawrence Zitnick},
  booktitle = {Proceedings of the European Conference on Computer Vision (ECCV)},
  year      = {2014},
  pages     = {740--755}
}

@InProceedings{Mandlekar2018,
  title     = {RoboTurk: A Crowdsourcing Platform for Robotic Skill Learning through Imitation},
  author    = {Ajay Mandlekar and Yuke Zhu and Animesh Garg and Jonathan Booher and Max Spero and Albert Tung and Julian Gao and John Emmons and Anchit Gupta and Emre Orbay and Silvio Savarese and Li Fei-Fei},
  booktitle = {Conference on Robot Learning (CoRL)},
  year      = {2018}
}

@InProceedings{Mirchandani2025RoboCrowd,
  title     = {RoboCrowd: Scaling Robot Data Collection Through Crowdsourcing},
  author    = {Suvir Mirchandani and D. D. Yuan and K. Burns and M. S. Islam and T. Z. Zhao and Chelsea Finn and Dorsa Sadigh},
  booktitle = {Proceedings of the IEEE International Conference on Robotics and Automation (ICRA)},
  year      = {2025}
}

@InProceedings{Mirchandani2025RoboCade,
  title     = {RoboCade: Gamifying Robot Data Collection},
  author    = {Suvir Mirchandani and M. Tang and J. Duan and J. I. Hamid and M. Cho and Dorsa Sadigh},
  booktitle = {Proceedings of the IEEE International Conference on Robotics and Automation (ICRA)},
  year      = {2026}
}

@InProceedings{Agarwal2026COBALT,
  title     = {{COBALT}: Crowdsourcing Robot Learning via Cloud-Based Teleoperation with Smartphones},
  author    = {Ayush Agarwal and Ansh Gandhi and Jeremy A. Collins and Omar Rayyan and Aryan Sarswat and Ranjani Koushik and Kaavya Menon and Masoud Moghani and Ajay Mandlekar and Animesh Garg},
  booktitle = {Proceedings of the IEEE International Conference on Robotics and Automation (ICRA)},
  year      = {2026}
}

@book{sutton2018reinforcement,
  title     = {Reinforcement Learning: An Introduction},
  author    = {Sutton, Richard S. and Barto, Andrew G.},
  edition   = {Second},
  year      = {2018},
  publisher = {MIT Press}
}

@article{asadi2017mean,
  title   = {Mean Actor Critic},
  author  = {Asadi, Kavosh and Allen, Cameron and Roderick, Melrose and Mohamed, Abdel-rahman and Konidaris, George and Littman, Michael},
  journal = {arXiv preprint arXiv:1709.00503},
  year    = {2017}
}

@inproceedings{ciosek2018expected,
  title     = {Expected Policy Gradients},
  author    = {Ciosek, Kamil and Whiteson, Shimon},
  booktitle = {Proceedings of the AAAI Conference on Artificial Intelligence},
  volume    = {32},
  number    = {1},
  year      = {2018}
}

@article{petit2019all,
  title   = {All-Action Policy Gradient Methods: A Numerical Integration Approach},
  author  = {Petit, Benjamin and Amdahl-Culleton, Loren and Liu, Yao and Smith, Jimmy and Bacon, Pierre-Luc},
  journal = {arXiv preprint arXiv:1910.09093},
  year    = {2019}
}

@inproceedings{kool2020estimating,
  title     = {Estimating Gradients and Variances with the Score Function via Sampling Without Replacement},
  author    = {Kool, Wouter and Van Hoof, Herke and Welling, Max},
  booktitle = {International Conference on Learning Representations (ICLR)},
  year      = {2020}
}

@inproceedings{nauman2023many,
  title        = {On Many-Actions Policy Gradient},
  author       = {Nauman, Michal and Cygan, Marek},
  booktitle    = {International Conference on Machine Learning (ICML)},
  pages        = {25769--25789},
  year         = {2023},
  organization = {PMLR}
}

@Article{Argall2009,
  title   = {A Survey of Robot Learning from Demonstration},
  author  = {Brenna D. Argall and Sonia Chernova and Manuela Veloso and Brett Browning},
  journal = {Robotics and Autonomous Systems},
  volume  = {57},
  number  = {5},
  pages   = {469--483},
  year    = {2009}
}

@InProceedings{Zhu2022VIOLA,
  title     = {{VIOLA}: Imitation Learning for Vision-Based Manipulation with Object Proposal Priors},
  author    = {Yifeng Zhu and Abhishek Joshi and Peter Stone and Yuke Zhu},
  booktitle = {Conference on Robot Learning (CoRL)},
  year      = {2022}
}

@InProceedings{Duan2023AR2D2,
  title     = {{AR2-D2}: Training a Robot Without a Robot},
  author    = {Jiafei Duan and Yi Ru Wang and Mohit Shridhar and Dieter Fox and Ranjay Krishna},
  booktitle = {Conference on Robot Learning (CoRL)},
  year      = {2023}
}

@InProceedings{Wang2024EVE,
  title     = {{EVE}: Enabling Anyone to Train Robots using Augmented Reality},
  author    = {Jun Wang and Chun-Cheng Chang and Jiafei Duan and Dieter Fox and Ranjay Krishna},
  booktitle = {ACM Symposium on User Interface Software and Technology (UIST)},
  year      = {2024}
}

@InProceedings{Wang2024DexCap,
  title     = {{DexCap}: Scalable and Portable Mocap Data Collection System for Dexterous Manipulation},
  author    = {Chen Wang and Haochen Shi and Weizhuo Wang and Ruohan Zhang and Li Fei-Fei and C. Karen Liu},
  booktitle = {Robotics: Science and Systems (RSS)},
  year      = {2024}
}

@InProceedings{Lepert2025Phantom,
  title     = {Phantom: Training Robots Without Robots Using Only Human Videos},
  author    = {Marion Lepert and Jiaying Fang and Jeannette Bohg},
  booktitle = {Conference on Robot Learning (CoRL)},
  year      = {2025}
}

@InProceedings{Bahl2022,
  title     = {Human-to-Robot Imitation in the Wild},
  author    = {Shikhar Bahl and Abhinav Gupta and Deepak Pathak},
  booktitle = {Robotics: Science and Systems (RSS)},
  year      = {2022}
}

@InProceedings{Qin2023AnyTeleop,
  title     = {{AnyTeleop}: A General Vision-Based Dexterous Robot Arm-Hand Teleoperation System},
  author    = {Yuzhe Qin and Wei Yang and Binghao Huang and Karl Van Wyk and Hao Su and Xiaolong Wang and Yu-Wei Chao and Dieter Fox},
  booktitle = {Robotics: Science and Systems (RSS)},
  year      = {2023}
}

@InProceedings{Wu2024GELLO,
  title     = {{GELLO}: A General, Low-Cost, and Intuitive Teleoperation Framework for Robot Manipulators},
  author    = {Philipp Wu and Alexander Sun and Jimmy Wu and Lirui Wang and Pieter Abbeel},
  booktitle = {IEEE/RSJ International Conference on Intelligent Robots and Systems (IROS)},
  year      = {2024}
}

@InProceedings{Zhao2023ALOHA,
  title     = {Learning Fine-Grained Bimanual Manipulation with Low-Cost Hardware},
  author    = {Tony Z. Zhao and Vikash Kumar and Sergey Levine and Chelsea Finn},
  booktitle = {Robotics: Science and Systems (RSS)},
  year      = {2023}
}

@InProceedings{Muller2024OpenBotFleet,
  title     = {{OpenBot-Fleet}: A System for Collective Learning with Real Robots},
  author    = {Matthias M{\"u}ller and Samarth Brahmbhatt and Ankur Deka and Quentin Leboutet and David Hafner and Vladlen Koltun},
  booktitle = {Proceedings of the IEEE International Conference on Robotics and Automation (ICRA)},
  year      = {2024}
}

@inproceedings{lepert2025masquerade,
  title     = {Masquerade: Learning from In-the-wild Human Videos using Data-Editing},
  author    = {Lepert, Marion and Fang, Jiaying and Bohg, Jeannette},
  booktitle = {Proceedings of the IEEE International Conference on Robotics and Automation (ICRA)},
  year      = {2026}
}

@article{vonAhn2008recaptcha,
  title     = {reCAPTCHA: Human-based character recognition via web security measures},
  author    = {von Ahn, Luis and Maurer, Benjamin and McMillen, Colin and Abraham, David and Blum, Manuel},
  journal   = {Science},
  volume    = {321},
  number    = {5895},
  pages     = {1465--1468},
  year      = {2008},
  publisher = {American Association for the Advancement of Science}
}

\clearpage
\appendix

\section{MAPS Data Reduces the Variance of the Imitation Learning Gradient Estimator.} \label{app:MAPS_variance_reduction_proof}
The following proof is adapted from \citep{nauman2023many}, modified for imitation learning, whose objective is to maximize the expected log-likelihood of expert actions under the learned policy $\pi_\theta$. Let $\rho_E(o)$ denote the observation visitation distribution of the expert dataset, and $\pi^*(a|o)$ denote the underlying multimodal expert policy. Let $g(o, a)$ be the gradient function evaluated at a specific observation $o$ and action $a$:
$$g(o,a) = \nabla_\theta \log \pi_\theta(a|o).$$
The standard gradient estimator $\hat{g}_1$ samples a single action per observation from the expert dataset: $\hat{g}_1 = g(o, a)$, where $o \sim \rho_E(o)$ and $a \sim \pi^*(a|o)$. By the Law of Total Variance, we can decompose the variance of this standard estimator by conditioning on the observation $o$:
$$\text{Var}(\hat{g}_1) = \mathbb{E}_{o \sim \rho_E}\left[\text{Var}_{a \sim \pi^*}(g(o,a) \mid o)\right] + \text{Var}_{o \sim \rho_E}(\mathbb{E}_{a \sim \pi^*}\left[g(o,a) \mid o\right]).$$
The two terms are:
\begin{enumerate}[leftmargin=*, topsep=-3pt, itemsep=0pt, parsep=0pt, partopsep=0pt]
    \item The action-dependent variance: $\mathbb{E}_{o}[\text{Var}_{a}(g(o,a) \mid o)]$, which arises from the stochasticity or multimodality of the valid expert actions at a given observation.
    \item The observation-dependent variance: $\text{Var}_{o}(\mathbb{E}_{a}[g(o,a) \mid o])$, which arises from the distribution of different states visited across trajectories.
\end{enumerate}
Consider now a MAPS dataset with width $K$. For a given observation $o$, we sample a set of $K$ independent valid actions $A = \{a^{(1)}, \dots, a^{(K)}\}$, where each $a^{(k)} \sim \pi^*(a|o)$. The MAPS gradient estimator $\hat{g}_K$ is the sample mean of the gradients for these actions:
$$\hat{g}_K = \frac{1}{K} \sum_{k=1}^K g(o, a^{(k)}).$$
Applying the Law of Total Variance to the MAPS estimator yields:
$$\text{Var}(\hat{g}_K) = \mathbb{E}_{o \sim \rho_E}[\text{Var}_{A}(\hat{g}_K \mid o)] + \text{Var}_{o \sim \rho_E}(\mathbb{E}_{A}[\hat{g}_K \mid o]).$$
By the linearity of expectation:
$$\mathbb{E}_{A}\left[ \frac{1}{K} \sum_{k=1}^K g(o, a^{(k)}) \mid o \right] = \mathbb{E}_{a \sim \pi^*}[g(o, a) \mid o],$$
so the observation-dependent variance term remains unchanged. However, because the $K$ actions in set $A$ are independent and identically distributed (i.i.d.) draws from $\pi^*(a|o)$, the variance of the sample mean conditioned on $o$ scales by $1/K$:
$$\text{Var}_{A}\left( \frac{1}{K} \sum_{k=1}^K g(o, a^{(k)}) \mid o \right) = \frac{1}{K^2} \sum_{k=1}^K \text{Var}_a(g(o, a^{(k)}) \mid o) = \frac{1}{K} \text{Var}_a(g(o, a) \mid o).$$
Substituting this back into the total variance equation for the MAPS estimator yields:
$$\text{Var}(\hat{g}_K) = \frac{1}{K} \mathbb{E}_{o \sim \rho_E}[\text{Var}_{a \sim \pi^*}(g(o,a) \mid o)] + \text{Var}_{o \sim \rho_E}(\mathbb{E}_{a \sim \pi^*}[g(o,a) \mid o]).$$
Finally, by subtracting the total variance of the MAPS estimator from the standard estimator, we obtain the exact theoretical variance reduction:
$$\Delta \text{Var} = \text{Var}(\hat{g}_1) - \text{Var}(\hat{g}_K) = \left( 1 - \frac{1}{K} \right) \mathbb{E}_{o \sim \rho_E}[\text{Var}_{a \sim \pi^*}(g(o,a) \mid o)].$$
Since $K > 1$ for a MAPS dataset, $( 1 - 1/K ) > 0$. Thus, as long as multimodality or variation exists in the underlying expert policy ($\text{Var}_a(g(o,a)|o) > 0$), utilizing MAPS data strictly and monotonically reduces the variance of the gradient estimator without introducing any bias. \hfill $\square$

\section{Crowdsourcing Details} \label{app:crowdsourcing_details}
\subsection{General Crowdsourcing Formulation} \label{app:general_crowdsourcing_formulation}
We provide here a general formulation of the crowdsourcing process, such that the formulation in the main text is an instantiation or a special case. In the general case, the crowdsourcing process involves the expansion of (1) a sequence of visited states $\mathcal{O}_{\text{visited}} = \smash{\{o_t\}_{t=0}^{T}}$ (the ``length'' of the dataset), and (2) the action sets $A_{t} = \smash{\{ a_{t}^{(k)} \}_{k=1}^{K_{t}}}$ associated with each state (the ``width'' of the dataset). A management system serves visited states to contributors based on a sampling criterion that prioritizes the frontier state $o_T$ (for trajectory expansion) or sparsely labeled past states (for density). When a contributor provides a candidate action for a state, it is classified as ``valid'' or ``invalid'' by an administrator to control data quality.

Independent of validity, we categorize actions based on their relationship with $\mathcal{O}_{\text{visited}}$. \textbf{Synchronous actions} ($\smash{A_t^{\text{sync}}}$) are collected while the robot is physically paused at the frontier state $o_t$, making them eligible for execution. From this set, the \textbf{executed action} $a_t^{\text{exec}} \in \smash{A_t^{\text{sync}}}$ is selected to transition the robot to $o_{t+1}$. Conversely, \textbf{asynchronous actions} ($\smash{A_t^{\text{async}}}$) are collected for $o_t$ after the robot has transitioned to $o_{t+1}$ or later. The full action set is $A_t = \smash{A_t^{\text{sync}}} \cup \smash{A_t^{\text{async}}},$ and the unexecuted set is $\smash{A_t^{\neg \text{exec}}} = A_t \setminus \{ a_t^{\text{exec}} \}$.

The system collects valid submissions for the frontier state $o_T$ until $|A_T^{\text{sync}}| \ge K_{\text{sync}}$, where $K_{\text{sync}}$ is the synchronous density target, at which point $a_T^{\text{exec}}\in A_T^{\text{sync}}$ is sampled and executed. Concurrently or otherwise, the system collects asynchronous actions for all prior states $o_{0 \dots T-1}$ until the asynchronous density target $K_{\text{async}}$ is met.

When the number of concurrent contributors $M$ significantly exceeds the synchronous density target ($M \gg K_{\text{sync}}$), $\mathcal{O}_{\text{visited}}$ expands with lower latency. However, demanding $M \gg K_{\text{sync}}$ while keeping $K_{\text{sync}}$ meaningfully large is infeasible in a research setting, thus we adopt in the main text the special case where $K_{\text{sync}} = 1$, $A_t^{\text{sync}} = \{a^{\text{exec}}_t\}$ for all $t$ and is supplied by the administrator.
\begin{figure}[h]
    \centering
    \includegraphics[width=\textwidth]{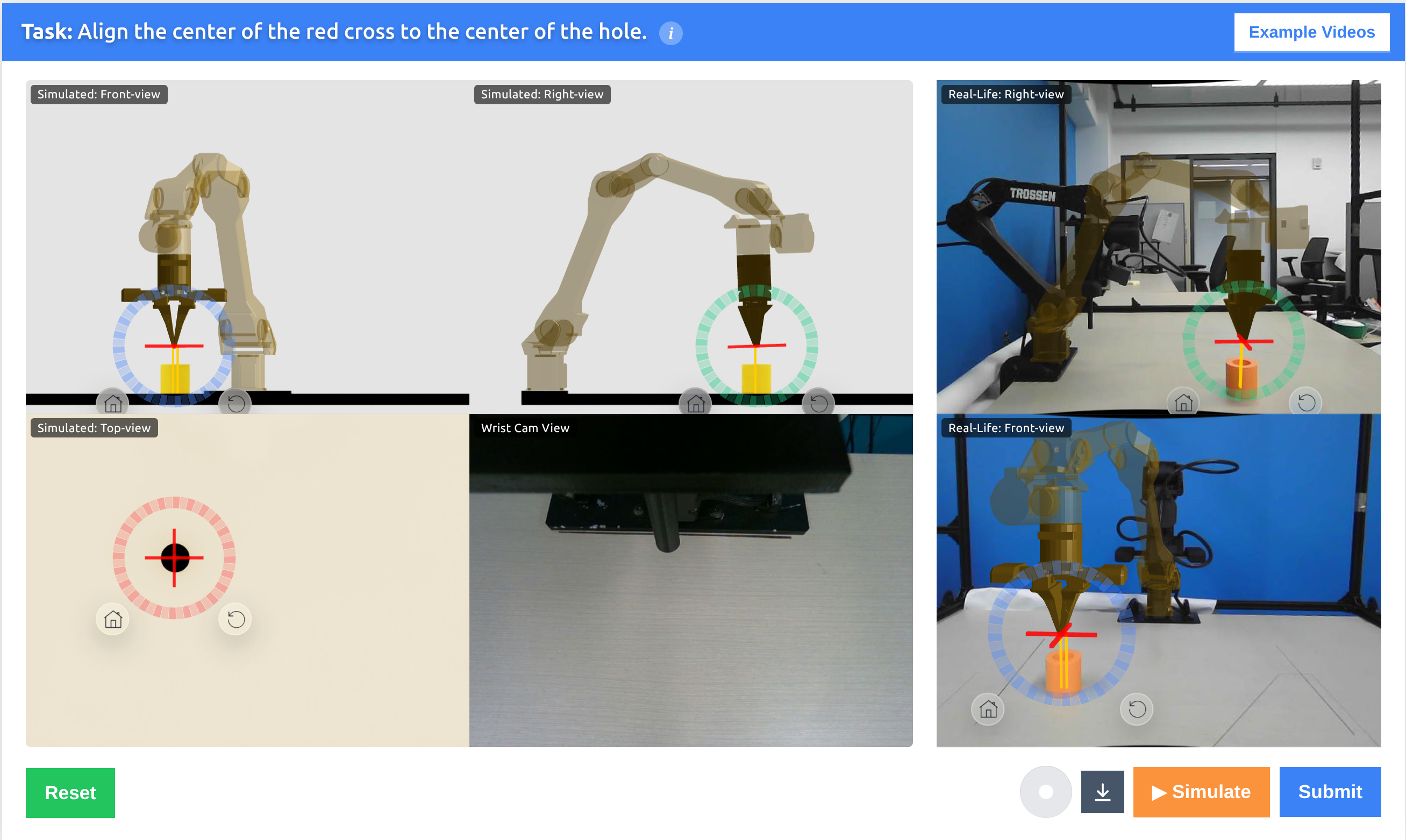}
    \caption{\textbf{Annotation User Interface.} Our annotation user interface contains a text description for the next keypose, six views, controls (a red cross that can be dragged to translate the end-effector, red, blue, and green wheels for rotating the end-effector), and other functionalities.}
    \label{fig:ui_insertion}
\end{figure}

\subsection{Crowdsourcing System and User-Interface} \label{app:crowdsourcing_ui}
The crowdsourcing system is implemented as a web application. The backend manages a pool of visited states, serves states whose action density target has not been fulfilled, renders robot motion and basic collisions with objects and fixtures in Isaac Lab when the user clicks the ``Simulate'' button, and records incoming submissions. The user-facing frontend contains a user-interface (UI) (see Fig.~\ref{fig:ui_insertion}) through which the users specify and submit next-goal-pose labels according to text and video prompts. Example videos are captured when the researcher teleoperates during the initial physical data collection phase and communicate expectations to the annotators.

For each task, we manually provide CAD models and text descriptions of the relevant objects. For each recorded keyframe, LangSAM segments the objects from the recorded observations using the text descriptions, and Any6D estimates their 6-DoF poses. The corresponding CAD meshes are then placed at the estimated poses in Isaac Lab to reconstruct the scene used for annotation. To reduce occasional pose-estimation errors, we run pose estimation multiple times offline and use the median estimated pose. Raw real-camera views of the recorded state are displayed alongside the reconstruction in the annotation interface.

\begin{figure}[h]
    \centering
    \includegraphics[width=\textwidth]{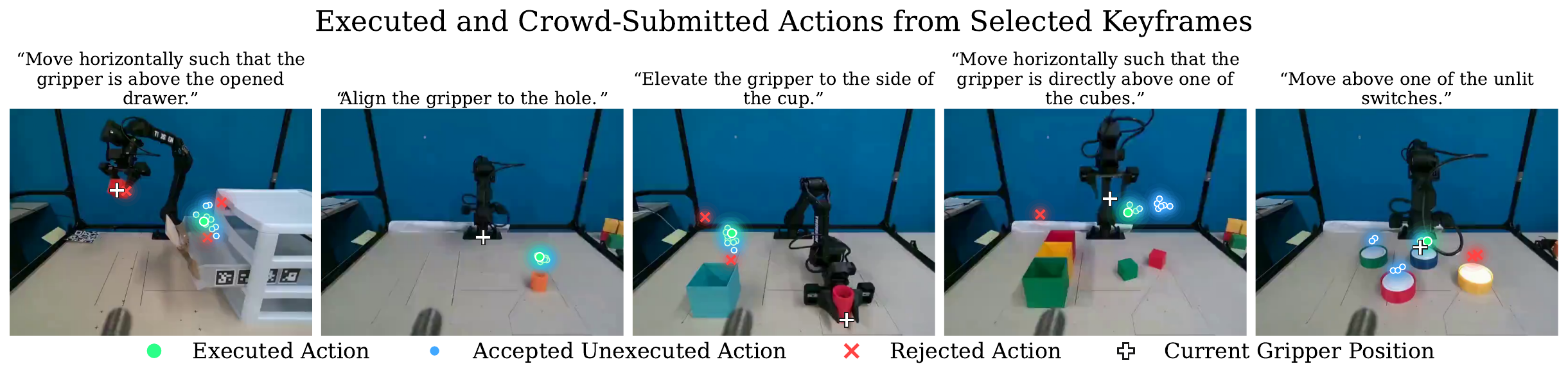}
    \caption{\textbf{Crowdsourced Labels for Selected Keyframes.} We show executed and crowd-submitted end-effector goal poses projected to the corresponding image observation for five keyframes, one from each task, from our crowdsourced datasets. Rejected actions can be semantically incorrect or physically incoherent. Accepted actions provide both intramodal and multimodal coverage of the valid action distribution.}
    \label{fig:example_validation}
\end{figure}

During crowdsourcing, the researcher uses an admin-facing interface to monitor progress and review submissions. During initial physical data collection, the researcher establishes a manipulation strategy for each keyframe variation of each task and follows that strategy during execution. Each crowdsourced submission is then validated according to three criteria: it must be consistent with the selected strategy, semantically aligned with the keyframe instructions, and physically safe and coherent. Validation is intended to remove off-strategy or implausible actions while preserving valid diversity within the strategy. For example, for the keyframe instruction ``Move above one of the unlit switches'' in Switches, the strategy is to place the end-effector at or around the center of any unlit puck light, with reasonable tolerance reflecting the size of the light. Across Drawer, Insertion, Pouring, Sorting, and Switches, the resulting submission acceptance rates are 51\%, 98\%, 66\%, 83\%, and 94\%, respectively. See examples of accepted and rejected submissions in Fig.~{\ref{fig:example_validation}}.

Our experiments were conducted on small, single-task datasets. We posit that for significantly larger and more diverse datasets, the submission validation criteria can be fundamentally relaxed: e.g., an annotation is accepted if it is safe, feasible, and progresses the task. We leave exploring this direction to future work.

\subsection{Crowdsourcing Process} \label{app:crowdsourcing_process}
\begin{figure*}[t]
    \centering
    \includegraphics[width=\textwidth]{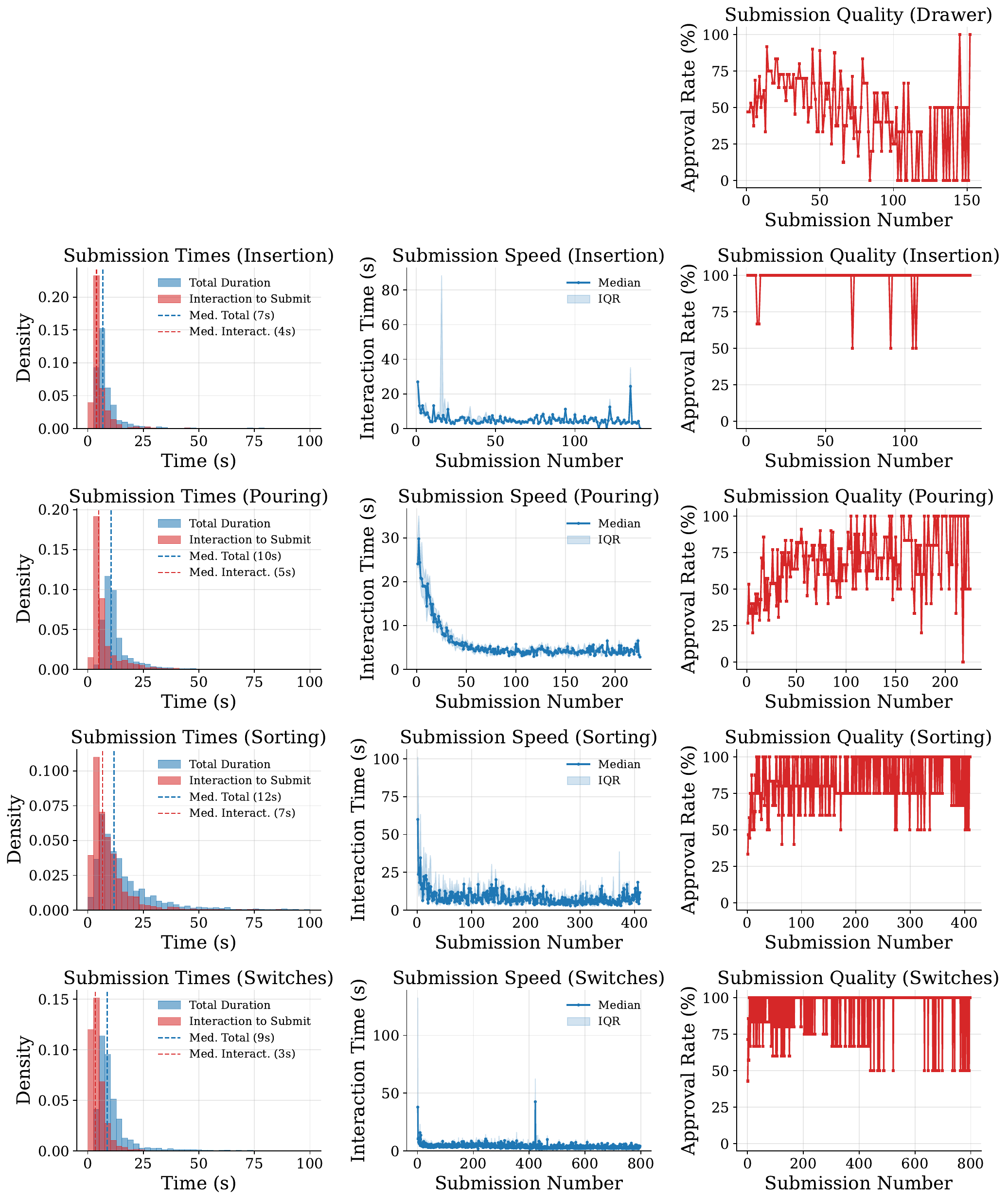}
    \caption{Crowdsourced annotation statistics across all five tasks (rows, alphabetical). \textbf{Left:} Distribution of submission times with median lines. \textbf{Center:} Median interaction time vs.\ per-user submission number with IQR shading. \textbf{Right:} Approval rate vs.\ per-user submission number. Drawer only shows quality (submission times were not recorded). Annotators generally improve in both speed and accuracy with practice.}
    \label{fig:crowdsourcing}
\end{figure*}

\begin{table}[htbp]
    \centering
    \caption{Participant Pool Statistics ($N=47$).}
    \label{tab:demographics}
    \begin{tabular}{@{}llcc@{}}
        \toprule
        \textbf{Category} & \textbf{Option} & \textbf{Count} & \textbf{Percentage} \\
        \midrule
        \multirow{2}{*}{Age} & 18--24 & 29 & 61.7\% \\
        & 25--34 & 18 & 38.3\% \\
        \midrule
        \multirow{2}{*}{Gender} & Male & 25 & 53.2\% \\
        & Female & 22 & 46.8\% \\
        \midrule
        \multirow{4}{*}{Current Status} & Graduate Student & 30 & 63.8\% \\
        & Undergraduate Student & 13 & 27.7\% \\
        & Other & 3 & 6.4\% \\
        & Staff & 1 & 2.1\% \\
        \midrule
        \multirow{4}{*}{General Robotics Exp.} & No experience & 25 & 53.2\% \\
        & Briefly learned & 14 & 29.8\% \\
        & Used infrequently & 7 & 14.9\% \\
        & Work with everyday & 1 & 2.1\% \\
        \midrule
        \multirow{3}{*}{Robot Arm Exp.} & No experience & 37 & 78.7\% \\
        & Briefly interacted & 9 & 19.1\% \\
        & Used infrequently & 1 & 2.1\% \\
        \bottomrule
    \end{tabular}
    \label{tab:participants}
\end{table}
We recruit a pool of 47 participants through university mailing lists and social media (see Table~\ref{tab:participants}), and conduct multiple rounds of crowdsourcing. As shown in Table~\ref{tab:participants}, most participants have little to no experience with robotics and robot arms. Submission time (measured by total duration: webpage refreshed to submission; and interaction duration: first UI interaction to submission), quality (measured by acceptance rate) and the progression of the two variables through each crowdsourcing round are shown in Fig.~\ref{fig:crowdsourcing}. Drawer (conducted first) only shows quality progression, as submission time was not logged. Overall, annotators improve in both speed and accuracy throughout the round, with Drawer being an exception. When crowdsourcing for Drawer, the system allowed annotators to refresh to change to a different keyframe variation (more about keyframes variations in~\ref{app:experiments}), the most difficult variations were thus left to the end of the session. Insertion and Switches only each have one variation (for Insertion: ``Align the center of the red cross to the center of the hole," shown in Fig.~\ref{fig:ui_insertion}), and are relatively easy to perform, resulting in the faster median submission time and higher submission acceptance rate from the beginning. Median total submission time is around 10 seconds across the four tasks with timing logs, making the crowdsourcing process highly flexible for each annotator compared to continuous teleoperation, which demands sustained operation for prolonged periods of time.

\section{Training Details} \label{app:training_details}
\subsection{Keypose Policy Configuration} \label{app:keypose_policy_details}
We implement the keypose policy by adapting Diffusion Policy in inference horizon, action-type dependent weighting in loss function, and inference scheme to learn to predict the next goal pose.
\paragraph{Model Architecture:}
The policy consists of two main modules: an observation encoder that processes raw sensory inputs into a global conditioning vector, and a 1D conditional U-Net that generates the action trajectory.

\textbf{1. Observation Encoder} \\
This module independently processes visual and state inputs before fusing them into a single representation.
\begin{description}
    \item[Image Preprocessing:] Images are resized so the shorter side is 96. A spatial crop of $84 \times 84$ is then applied (random crop during training; center crop during evaluation).
    \item[Vision Backbone:] Cropped images are processed by a ResNet-18 trained from scratch. Batch Normalization is replaced with Group Normalization using a group size of \texttt{feature\_dim // 16} (16 channels per group).
    \item[Spatial Softmax:] A spatial softmax layer extracts 32 2D coordinate keypoints from the backbone's feature maps. This yields a $64$-dimensional ($32 \times 2$) visual feature vector per camera.
    \item[State Fusion:] Proprioceptive state vectors are normalized and concatenated directly with the visual feature vectors to produce the final global conditioning vector.
\end{description}

\textbf{2. 1D Denoising U-Net} \\
The U-Net predicts the noise to be removed from the action sequence, conditioned on the observation history and the diffusion timestep.
\begin{description}
    \item[Temporal Structure:] The model utilizes an observation history of $N_{\text{obs\_steps}} = 2$ and a prediction horizon of $H = 2$, executing $N_{\text{action\_steps}} = 1$ action step per inference cycle.
    \item[Network Topology:] The architecture features down-sampling layers with channel dimensions of (512, 1024). The core convolutional blocks consist of a 1D Convolution (kernel size 5) $\rightarrow$ Group Normalization (8 groups) $\rightarrow$ Mish activation.
    \item[Conditioning Mechanism (FiLM):] A 128-dimensional sinusoidal timestep embedding is mapped through a 2-layer MLP (Mish activation). This timestep embedding is concatenated with the global conditioning vector (flattened across $N_{\text{obs\_steps}}$) and applied to the U-Net residual blocks via Feature-wise Linear Modulation (FiLM).
\end{description}

The global conditioning vector (concatenated observation embeddings across $N_{\text{obs\_steps}}$) is concatenated with the timestep embedding.

\paragraph{Hyperparameters}
\begin{center}
\begin{tabular}{ll}
\hline
\textbf{Diffusion and Denoising Parameters} &  \\ \hline
Noise Scheduler & DDPM (Denoising Diffusion Probabilistic Model) \\
Train Timesteps ($T$) & 100 \\
Beta Schedule & \texttt{squaredcos\_cap\_v2} \\
Beta Start ($\beta_{\text{start}}$) & 0.0001 \\
Beta End ($\beta_{\text{end}}$) & 0.02 \\
Prediction Type & epsilon/noise prediction \\
Sample Clipping & Enabled, range $[-1.0, 1.0]$ \\ \hline
\end{tabular}
\end{center}

\begin{center}
\begin{tabular}{ll}
\hline
\textbf{Optimization and Training Parameters} & \\ \hline
Optimizer & Adam \\
Learning Rate & $1 \times 10^{-4}$ \\
Betas ($\beta_1, \beta_2$) & $(0.95, 0.999)$ \\
Epsilon ($\epsilon$) & $1 \times 10^{-8}$ \\
Weight Decay & $1 \times 10^{-6}$ \\
Learning Rate Scheduler & Cosine decay with linear warmup \\
Warmup Steps & 500 \\ \hline
\end{tabular}
\end{center}

\paragraph{Action type weighting:}
During training, the policy weighs the mean squared error (MSE) loss based on the action type of the batch sample:
\begin{equation}
\mathcal{L} = \frac{\sum_{i=1}^B w_i \cdot \text{MSE}(\epsilon_i, \hat{\epsilon}_i)}{\sum_{i=1}^B w_i},
\end{equation}
where $w_i$ is the sample weight, and the MSE for a single sample $i$ is defined as:
\begin{equation}
\text{MSE}(\epsilon_i, \hat{\epsilon}_i) = \frac{1}{H \cdot D} \sum_{h=1}^H \sum_{d=1}^D (\epsilon_{ihd} - \hat{\epsilon}_{ihd})^2,
\end{equation}
where $H$ is the sample horizon, and $D$ is the action dimension. As introduced previously, \textbf{executed keyframe actions} ($t \in \mathcal{K}, g = g^{\text{exec}}_t$) receive weight $w_{\text{key, exec}}$; \textbf{unexecuted keyframe actions} ($t \in \mathcal{K}$, $g \in G_t^{\neg \text{exec}}$) are weighted by $w_{\text{key, } \neg \text{exec}}$; \textbf{intermediate actions} ($t \notin \mathcal{K}$) are assigned $w_{\text{inter}}$; \textbf{single-mode keyframe actions} are weighted by $w_{\text{single}}$. We hold the total weight of keypose frames to be constant to ensure fair comparisons between $K=1,2,5,10$ (i.e., $w_{\text{key, exec}} + (K-1) \cdot w_{\text{key, } \neg \text{exec}} = 65.0$).

\begin{table}[h!]
\centering
\begin{tabular}{lcccc}
\hline
\textbf{$K$} & \textbf{$w_{\text{inter}}$} & \textbf{$w_{\text{key, exec}}$} & \textbf{$w_{\text{key, } \neg \text{exec}}$} & \textbf{$w_{\text{single}}$} \\ \hline
$1$  & $1.0$ & $65.0$ & $5.0$ & $65.0$ \\
$2$  & $1.0$ & $60.0$ & $5.0$ & $65.0$ \\
$5$  & $1.0$ & $45.0$ & $5.0$ & $65.0$ \\
$10$ & $1.0$ & $20.0$ & $5.0$ & $65.0$ \\ \hline
\end{tabular}
\end{table}

\paragraph{Inference Scheme:}
To interface with the robot hardware and control loops, the policy employs a dynamic queue-management system. This scheme guarantees responsive execution while preventing the robot from stalling or executing outdated plans.

\begin{description}
    \item[Memory Buffers:] The control loop maintains a rolling history of the past $N_{\text{obs\_steps}} = 2$ observations and the newest generated goal pose.
    \item[Pause-Based Replanning:] When we reach the current generated goal pose, the robot would pause, and it is time to generate a new goal pose (replan). For pause detection, the control loop computes a moving average of the last 5 joint states. If the maximum absolute joint difference from this average is $< 0.001$ rad and the latest joint delta is $< 0.0002$ rad, the robot is classified as paused. The current generated goal pose is cleared and the policy is invoked to generate a new goal pose.
    \item[Time-Based Replanning:] To implement closed-loop control grounded in current visual feedback, the control loop clears the current goal pose and invokes the policy for new goal generation every 10 iterations.
    \item[Diffusion Generation:] When the current goal pose is \texttt{None} , the 1D Conditional U-Net performs 100 reverse diffusion steps to predict the next goal pose.
\end{description}

\subsection{Trajectory Policy Configuration} \label{app:trajectory_policy_details}
Our trajectory policy setup follows the standard Diffusion Policy implementation but adapted the data loader to support hybrid-horizon co-training, enabling the model to learn simultaneously from continuous expert trajectories and multi-action crowdsourced keypose data.

To adapt the single-step goals $g_t$ from $\mathcal{D}_{\text{MAPS}}$ to the sequence horizon $H = 16$ required by the trajectory policy, we synthesize action chunks $\mathbf{a}_t^{\text{MAPS}}$ via linear interpolation. Specifically, the trajectory is interpolated from the current end-effector pose $g_t^{\text{curr}}$ to the target keypose goal $g_t$ at a constant rate determined by the average velocity $\bar{v}_{\text{batch}}$ computed from the expert demonstrations $\mathcal{D}_{\text{traj}}$ in the current training batch. If the target goal is reached prior to step $H$, the remaining steps of the chunk are padded by clamping to the goal pose $g_t$.

\paragraph{Model Architecture:}
The policy consists of two main modules: an Observation Encoder that processes raw sensory inputs into a global conditioning vector, and a 1D Conditional U-Net that generates the action trajectory.

\textbf{1. Observation Encoder} \\
This module independently processes visual and state inputs before fusing them into a single representation.
\begin{description}
    \item[Image Preprocessing:] Images are resized so the shorter side is 96. A spatial crop of $84 \times 84$ is then applied (random crop during training; center crop during evaluation).
    \item[Vision Backbone:] Cropped images are processed by a ResNet-18 trained from scratch. Batch Normalization is replaced with Group Normalization using a group size of \texttt{feature\_dim // 16} (16 channels per group).
    \item[Spatial Softmax:] A spatial softmax layer extracts 32 2D coordinate keypoints from the backbone's feature maps. This yields a $64$-dimensional ($32 \times 2$) visual feature vector per camera.
    \item[Dataset ID Conditioning:] A 2-dimensional one-hot vector representing the data source (\texttt{[1, 0]} for expert trajectories, \texttt{[0, 1]} for crowdsourced keypose demonstrations) is appended to the conditioning vector. We find that using this conditioning brings no significant performance gain in our training settings, and therefore trained and evaluated all trajectory policies with a constant 0 conditioning, ignoring dataset ID.
    \item[State Fusion:] Proprioceptive state vectors are normalized and concatenated directly with the visual feature and dataset ID vectors to produce the final global conditioning vector.
\end{description}

\textbf{2. 1D Denoising U-Net} \\
The U-Net predicts the noise to be removed from the action sequence, conditioned on the observation history and the diffusion timestep.
\begin{description}
    \item[Temporal Structure:] The model utilizes an observation history of $N_{\text{obs\_steps}} = 2$ and a prediction horizon of $H = 16$, executing $N_{\text{action\_steps}} = 8$ action steps per inference cycle.
    \item[Network Topology:] The architecture features down-sampling layers with channel dimensions of (512, 1024, 2048) across 3 stages. The core convolutional blocks consist of a 1D Convolution (kernel size 5) $\rightarrow$ Group Normalization (8 groups) $\rightarrow$ Mish activation.
    \item[Conditioning Mechanism (FiLM):] A 128-dimensional sinusoidal timestep embedding is mapped through a 2-layer MLP (Mish activation). This timestep embedding is concatenated with the global conditioning vector (flattened across $N_{\text{obs\_steps}}$) and applied to the U-Net residual blocks via Feature-wise Linear Modulation (FiLM).
\end{description}

\paragraph{Hyperparameters}
\begin{center}
\begin{tabular}{ll}
\hline
\textbf{Diffusion and Denoising Parameters} &  \\ \hline
Noise Scheduler & DDPM (Denoising Diffusion Probabilistic Model) \\
Train Timesteps ($T$) & 100 \\
Beta Schedule & \texttt{squaredcos\_cap\_v2} \\
Beta Start ($\beta_{\text{start}}$) & 0.0001 \\
Beta End ($\beta_{\text{end}}$) & 0.02 \\
Prediction Type & epsilon/noise prediction \\
Sample Clipping & Enabled, range $[-1.0, 1.0]$ \\ \hline
\end{tabular}
\end{center}

\begin{center}
\begin{tabular}{ll}
\hline
\textbf{Optimization and Training Parameters} & \\ \hline
Optimizer & Adam \\
Learning Rate & $1 \times 10^{-4}$ \\
Betas ($\beta_1, \beta_2$) & $(0.95, 0.999)$ \\
Epsilon ($\epsilon$) & $1 \times 10^{-8}$ \\
Weight Decay & $1 \times 10^{-6}$ \\
Learning Rate Scheduler & Cosine decay with linear warmup \\
Warmup Steps & 500 \\ \hline
\end{tabular}
\end{center}

\paragraph{Action type weighting.}
During training, the policy weighs the mean squared error (MSE) loss based on the action type of the batch sample:

\begin{equation}
\mathcal{L} = \frac{\sum_{i=1}^B w_i \cdot \text{MSE}(\epsilon_i, \hat{\epsilon}_i)}{\sum_{i=1}^B w_i},
\end{equation}
where $w_i$ is the sample weight. Under the hybrid training structure, normal expert trajectories receive a weight of $w_{\text{traj}} = 130.0$. For the crowdsourced data portion: executed keyframe actions ($t \in \mathcal{K}, g = g^{\text{exec}}_t$) receive weight $w_{\text{key, exec}}$; unexecuted keyframe actions ($t \in \mathcal{K}$, $g \in G_t^{\neg \text{exec}}$) are weighted by $w_{\text{key, } \neg \text{exec}} = 5.0$; intermediate actions ($t \notin \mathcal{K}$) are assigned $w_{\text{inter}} = 1.0$; single-mode keyframe actions are weighted by $w_{\text{single}} = 65.0$. The total weight of keypose frames is conserved to be constant across different crowd sizes $k$ ($w_{\text{key, exec}} + (K-1) \cdot w_{\text{key, } \neg \text{exec}} = 65.0$).

\begin{table}[h!]
\centering
\begin{tabular}{lccccc}
\hline
\textbf{$K$} & \textbf{$w_{\text{inter}}$} & \textbf{$w_{\text{key, exec}}$} & \textbf{$w_{\text{key, } \neg \text{exec}}$} & \textbf{$w_{\text{single}}$} & \textbf{$w_{\text{traj}}$} \\ \hline
$1$  & $1.0$ & $65.0$ & $5.0$ & $65.0$ & $130.0$ \\
$2$  & $1.0$ & $60.0$ & $5.0$ & $65.0$ & $130.0$ \\
$5$  & $1.0$ & $45.0$ & $5.0$ & $65.0$ & $130.0$ \\
$10$ & $1.0$ & $20.0$ & $5.0$ & $65.0$ & $130.0$ \\ \hline
\end{tabular}
\end{table}

\paragraph{Inference Scheme:}
To interface with the robot hardware, the Trajectory Policy employs a standard open-loop execution scheme:

\begin{description}
    \item[Memory Buffers:] The control loop maintains a rolling history of the past $N_{\text{obs\_steps}} = 2$ observations and queues the predicted action trajectory.
    \item[Nominal Execution:] The control loop sequentially executes $N_{\text{action\_steps}} = 8$ steps from the queued trajectory before triggering the next replanning cycle.
    \item[Diffusion Generation:] When the action queue is empty, the 1D Conditional U-Net performs 100 reverse diffusion steps to generate the next action trajectory of horizon $H=16$.
\end{description}

\section{Additional Experiment Details and Results} \label{app:experiments}
\subsection{Hardware and Environment Setup}
We set up a Trossen WidowX AI arm, an Intel RealSense D455 facing the robot and the workspace from the front, and an Intel RealSense D405 as the wrist camera. For continuous teleoperation, we use a second WidowX AI arm to form a leader-follower setup. For the real-life views in the crowdsourcing UI, we set up two additional webcams, one facing the robot from the front, one from the side.

\subsection{Tasks Setup}
We describe each tested task in detail. For each task, we design a progression of keyframes, a strategy to follow, during data collection and crowdsourcing. The keyframe types are \textbf{single mode} and \textbf{variety}. Single mode keyframes have next-step instructions such as close or open the gripper, return to the home position, etc., where the valid action distribution is a dirac delta function, and additional action labels would not be useful. These instructions have a bracketed action preset printed after them (e.g. [HOME]). Variety keyframes on the contrary admit crowd annotations.
\subsubsection{Task: Drawer}
\textbf{Keyframes:}\\
\includegraphics[width=\textwidth]{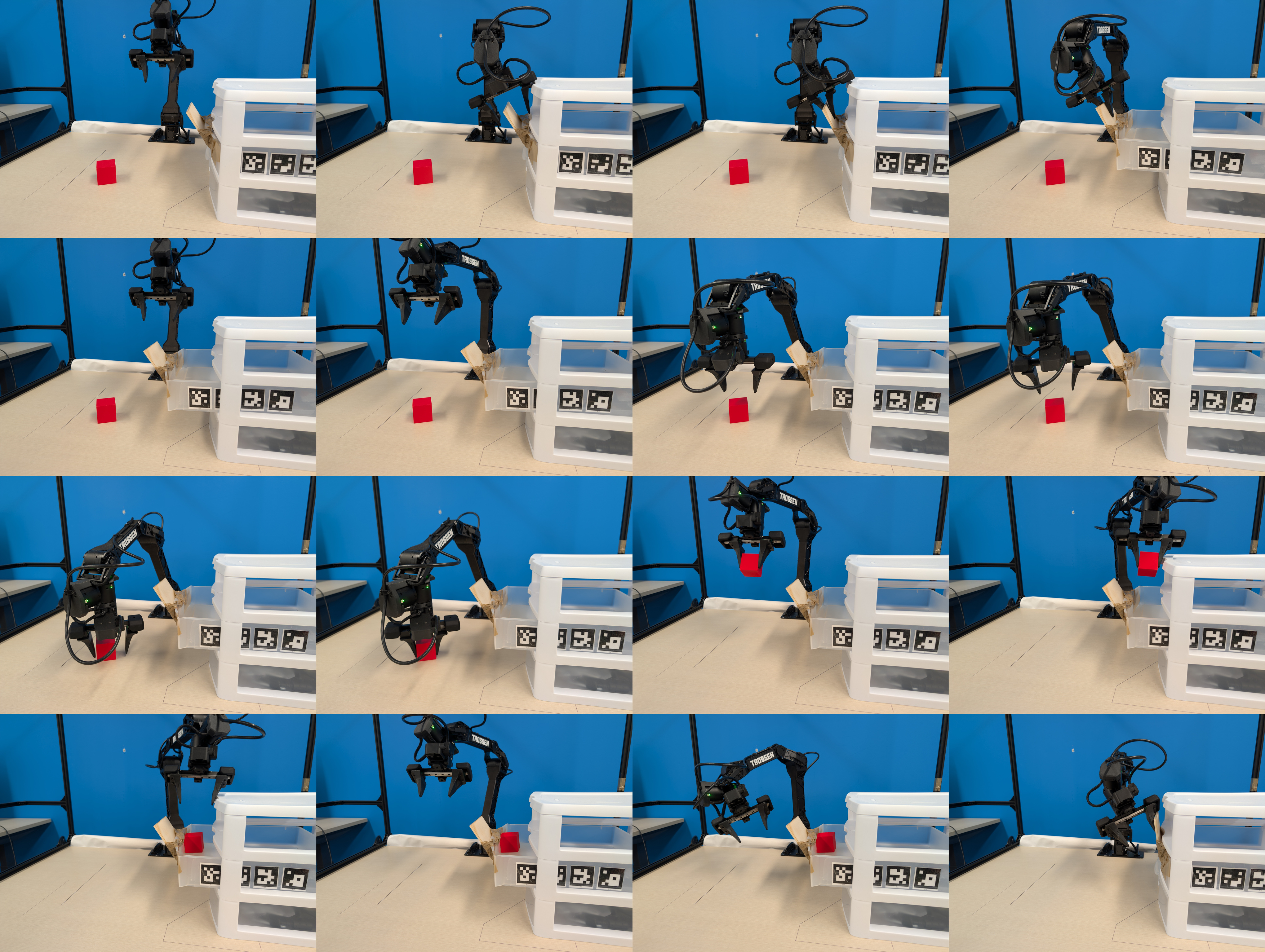}
\textbf{Objects:} Drawer, with a plate taped to the middle layer for the robot to grasp. One red cube with 5cm edges.\\
\textbf{End Effector:} Parallel gripper.\\
\textbf{Initialization:} The cube is initialized along a 30cm line, with a vertex or a side touching the line and random orientation. The position of the drawer is fixed.\\
\textbf{Keyframe Instructions:}
\begin{enumerate}[topsep=-3pt, itemsep=0pt, parsep=0pt, partopsep=0pt]
    \item Position the arm to grip the handle of the middle drawer.
    \item Grasp the drawer handle. [GRASP]
    \item Pull the drawer open.
    \item Release the gripper. [RELEASE]
    \item Return to the home position. [HOME]
    \item Move horizontally such that the gripper is in the general area above the cube.
    \item Lower and pitch the gripper till vertical to be $\sim$3cm above the cube.
    \item Rotate the gripper to align with grasp angle
    \item Lower the gripper to prepare for grasp.
    \item Close gripper to grasp. [GRASP]
    \item Raise the gripper such that the gripper and the cube clear the height of the middle drawer's handle.
    \item Move horizontally such that the gripper is above the opened drawer.
    \item Open gripper to release object into the drawer. [RELEASE]
    \item Move horizontally left to be slightly left of the drawer.
    \item Move down the gripper to be on the same level as the drawer, and tilt to the left, to prepare to push the drawer close.
    \item Push the drawer close.
    \item End. [END]
\end{enumerate}
\textbf{Evaluation Success Criteria:} the middle layer of the drawer is closed with the red cube inside.
\subsubsection{Task: Insertion}
\textbf{Keyframes:}\\
\includegraphics[width=\textwidth]{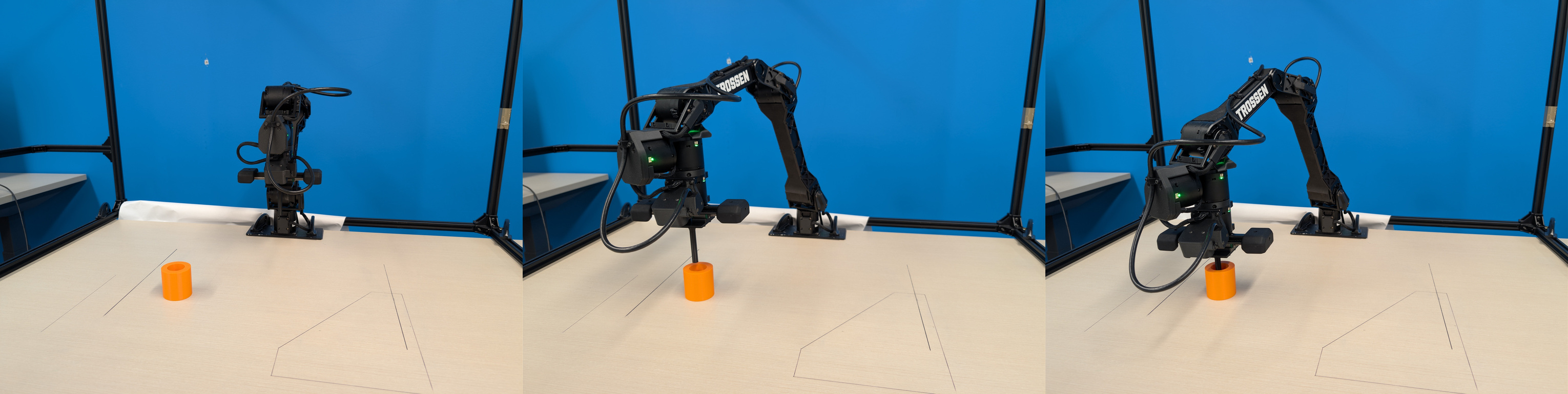}
\textbf{Objects:} One hollow orange cylinder, with 3cm diameter opening.\\
\textbf{End Effector:} A 7cm long, 1cm diameter pole.\\
\textbf{Initialization:} The cylinder is initialized within a 30cm by 70cm rectangular region on the desk.\\
\textbf{Keyframe Instructions:}
\begin{enumerate}[topsep=-3pt, itemsep=0pt, parsep=0pt, partopsep=0pt]
    \item Align the center of the red cross to the center of the hole.
    \item Insert. [DOWN]
    \item End. [END]
\end{enumerate}
\textbf{Evaluation Success Criteria:} The end-effector inserts into the hole without touching the top surface. Also success if end-effector touches, but lifts up before inserting without touching.
\subsubsection{Task: Pouring}
\textbf{Keyframes:}\\
\includegraphics[width=\textwidth]{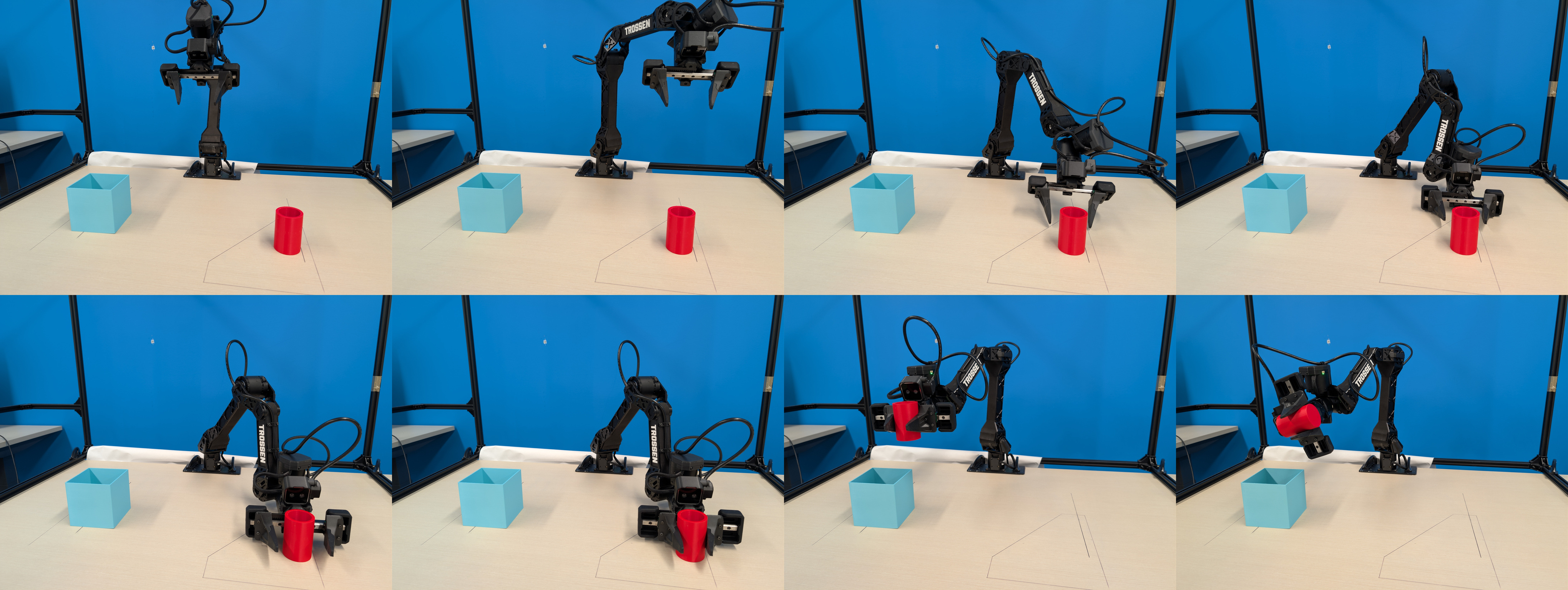}
\textbf{Objects:} One 3cm diameter red cylindrical cup. Two dozen nails. One cyan cubic container with 10cm edges.\\
\textbf{End Effector:} Parallel gripper.\\
\textbf{Initialization:} The cup is initialized within a triangular area on the desk (visibly drawn on the desk in the images). The container is within a 30cm by 30cm rectangular region on the left side of the desk.\\
\textbf{Keyframe Instructions:}
\begin{enumerate}[topsep=-3pt, itemsep=0pt, parsep=0pt, partopsep=0pt]
    \item Move the arm to be over and behind the cup.
    \item Lower the gripper vertically to the bottom.
    \item Flatten the gripper. [FLATTEN]
    \item Push forward to prepare for a deep grip of the cup.
    \item Close the gripper. [GRASP]
    \item Elevate the gripper to the side of the cup.
    \item Pour.
    \item End. [END]
\end{enumerate}
\textbf{Evaluation Success Criteria:} All nails besides at most three are poured into the container. If nails get stuck due to friction but the motion is correct, also success.
\subsubsection{Task: Sorting}
\textbf{Keyframes:}\\
\includegraphics[width=\textwidth]{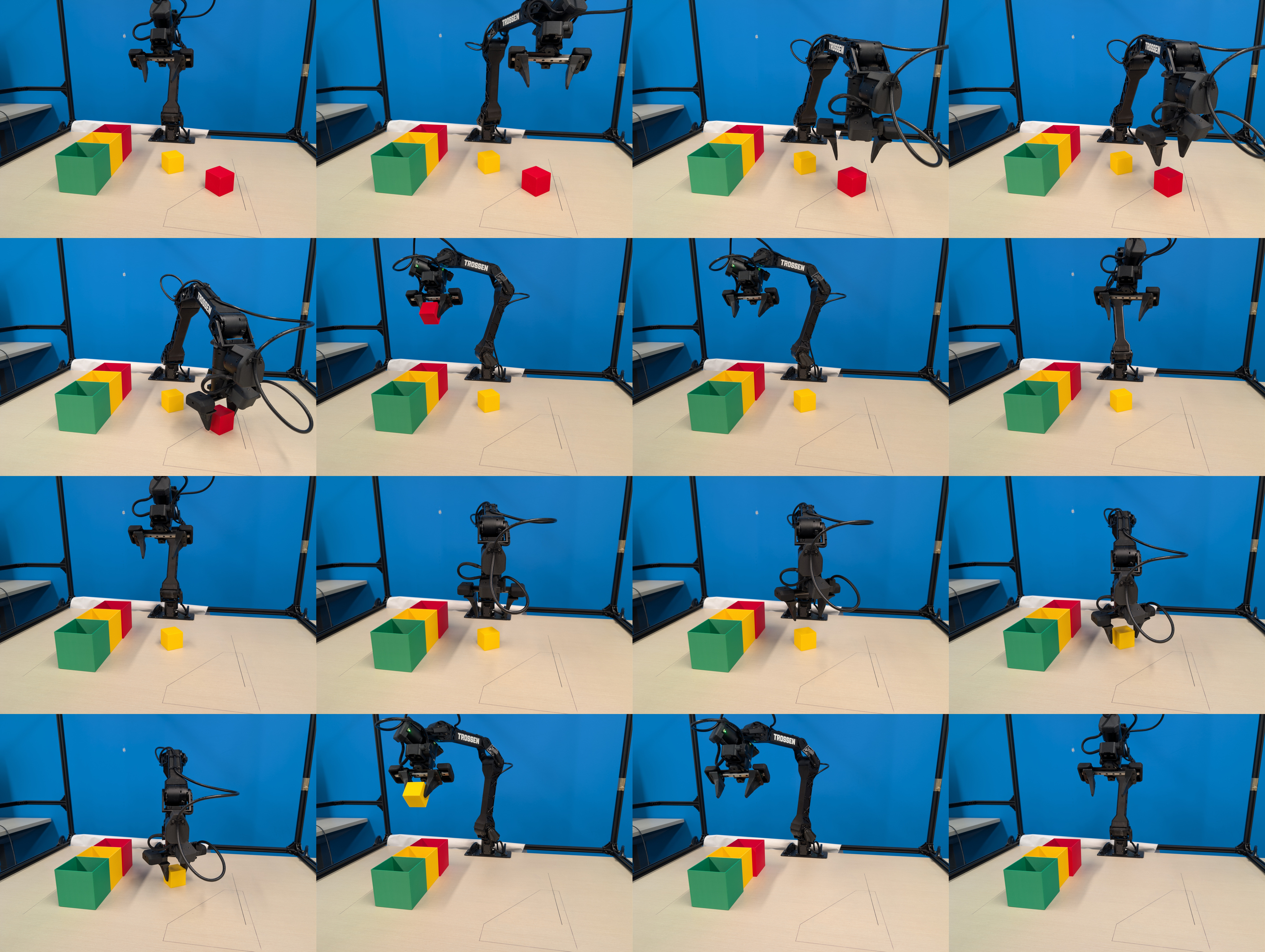}
\textbf{Objects:} Three 5cm cubes (red, yellow, green). Three 10cm containers (red, yellow, green).\\
\textbf{End Effector:} Parallel gripper.\\
\textbf{Initialization:} Two of the three cubes are initialized within a 30cm by 40cm rectangular region on the desk. The containers are stationary.\\
\textbf{Keyframe Instructions:}
\begin{enumerate}[topsep=-3pt, itemsep=0pt, parsep=0pt, partopsep=0pt]
    \item Move horizontally such that the gripper is directly above one of the cubes.
    \item Lower and pitch the gripper till vertical to be $\sim$4cm above the cube.
    \item Rotate the gripper to align with grasp angle.
    \item Lower the gripper to prepare for grasp.
    \item Close gripper to grasp. [GRASP]
    \item Raise the cube to be above the corresponding container [RED/YELLOW/GREEN]
    \item Open gripper to release object into the container. [RELEASE]
    \item Return to Home position. [HOME]
    \item Move horizontally such that the gripper is directly above one of the cubes.
    \item Lower and pitch the gripper till vertical to be $\sim$3cm above the cube.
    \item Rotate the gripper to align with grasp angle.
    \item Lower the gripper to prepare for grasp.
    \item Close gripper to grasp. [GRASP]
    \item Raise the cube to be above the corresponding container [RED/YELLOW/GREEN]
    \item Open gripper to release object into the container. [RELEASE]
    \item End. [END]
\end{enumerate}
\textbf{Evaluation Success Criteria:} All cubes are inside the container of the corresponding color.
\subsubsection{Task: Switches}
\textbf{Keyframes:}\\
\includegraphics[width=\textwidth]{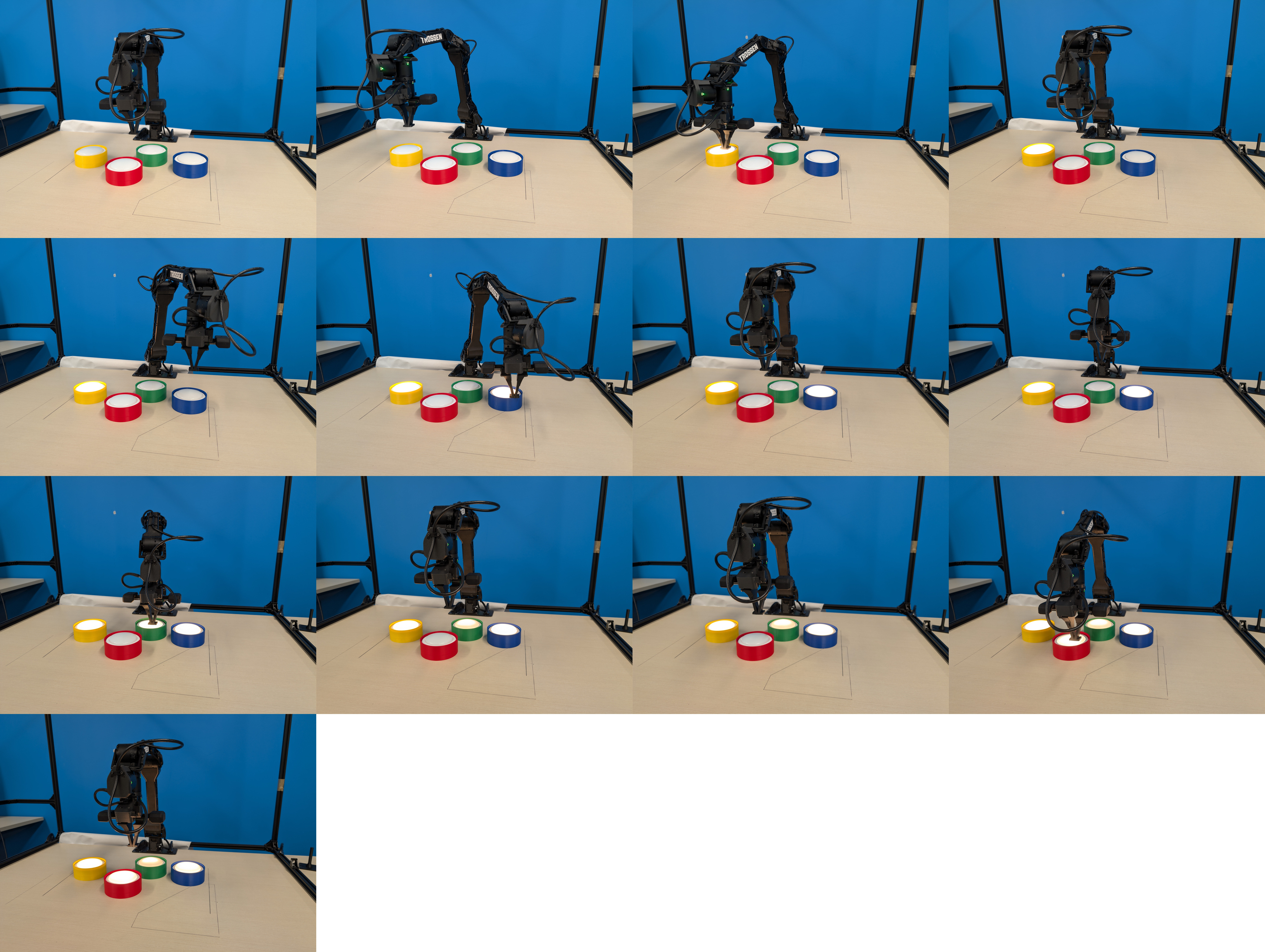}
\textbf{Objects:} Four puck lights, each of which turns on and off upon pressing, placed inside four cylindrical sleeves (red, green, yellow, blue).\\
\textbf{End Effector:} Modified parallel gripper, with a flat tip and reinforced for pressing.\\
\textbf{Initialization:} The puck lights are initialized within a 30cm by 70cm rectangular region on the desk.\\
\textbf{Keyframe Instructions:}
\begin{enumerate}[topsep=-3pt, itemsep=0pt, parsep=0pt, partopsep=0pt]
    \item Move above one of the unlit switches.
    \item Move down to press the switch. [DOWN]
    \item Return to Home position. [HOME]
    \item Move above one of the unlit switches.
    \item Move down to press the switch. [DOWN]
    \item Return to Home position. [HOME]
    \item Move above one of the unlit switches.
    \item Move down to press the switch. [DOWN]
    \item Return to Home position. [HOME]
    \item Move above one of the unlit switches.
    \item Move down to press the switch. [DOWN]
    \item Return to Home position. [HOME]
    \item End. [END]
\end{enumerate}
\textbf{Evaluation Success Criteria:} Four switches lit up.
\textbf{Note on Switches:} This task is to the greatest extent non-Markovian among all tested tasks, especially when the policy ingests a small number of past observations, due to how much the path of the end-effector intersects itself. We therefore devise Switches$^-$ as a simplified version (although still challenging, evident in the number of episodes needed to attain higher success rates in the keypose policy setting) to avoid all evaluated models collapsing to very low success rates. Due to the same reason, when evaluating Switches$^+$, we reset the robot to the home position after the robot successfully presses each light. This is done consistently across all Switches$^+$ configurations.

\subsection{What Drives the Gain from Wider MAPS?}
\label{app:gain_analysis}

We hypothesize that the gains from increasing MAPS width arise from two compounding effects: \emph{within-mode densification} and \emph{multimodal coverage}. Within-mode densification provides a denser sampling of locally valid actions around a particular solution, while multimodal coverage increases the probability that distinct valid action modes are represented at the same state.

To clarify a common misunderstanding, these gains cannot be attributed to broadening the observation distribution around the demonstrated trajectories. As discussed in Sec.~3.3, CrowdMAPS adds alternative action labels only at already visited observations and does not add successor observations induced by those alternatives ($p_{\mathcal{D}_{\mathrm{MAPS}}}(o) = p_{\mathcal{D}_{\mathrm{exec}}}(o)$). Thus, CrowdMAPS provides no additional recovery supervision at off-trajectory states; the observed gains must arise from changes to the action supervision at the original visited states.

\begin{figure}[t]
    \centering
    \begin{minipage}[t]{0.49\linewidth}
        \centering
        \includegraphics[
            height=0.20\textheight,
            width=\linewidth,
            keepaspectratio
        ]{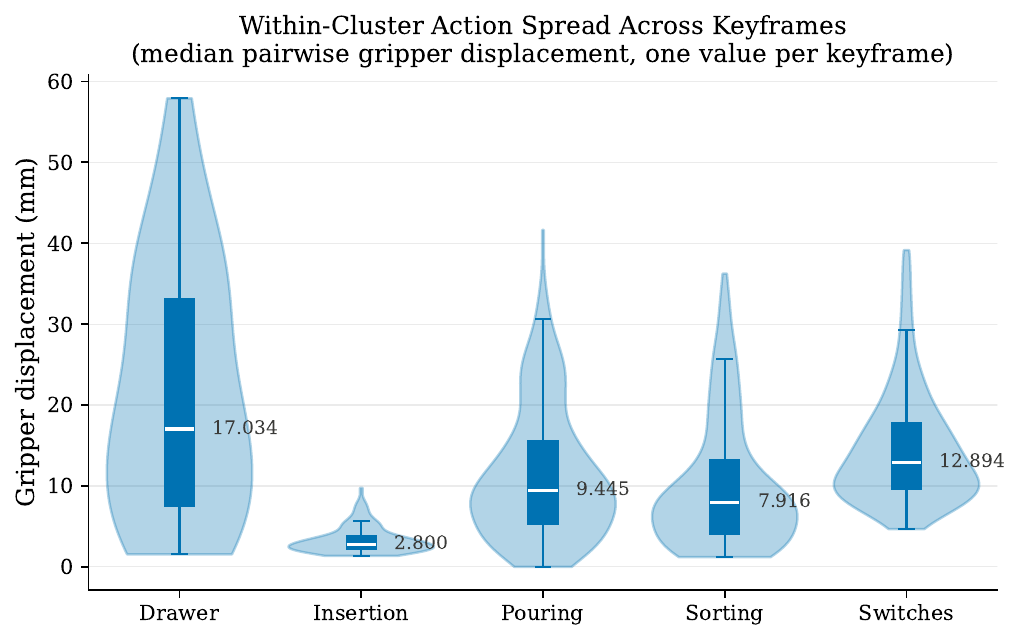}
        \small (a) Within-mode spread
    \end{minipage}
    \hfill
    \begin{minipage}[t]{0.49\linewidth}
        \centering
        \includegraphics[
            height=0.20\textheight,
            width=\linewidth,
            keepaspectratio
        ]{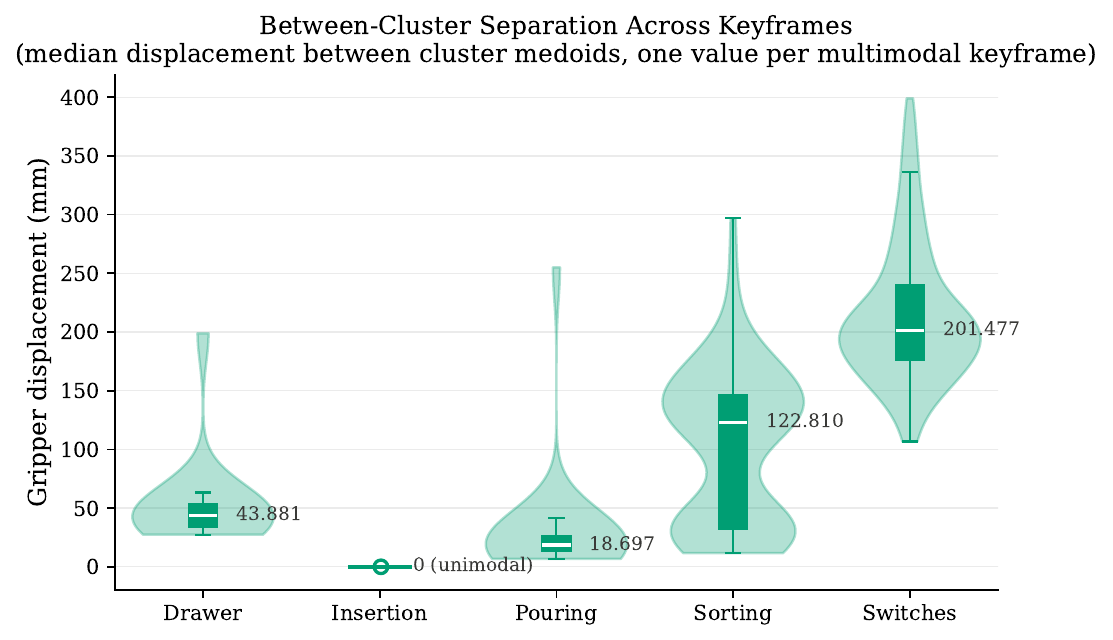}
        \small (b) Between-mode separation
    \end{minipage}
    \caption{
    \textbf{Geometric structure of crowdsourced actions.}
    (a) Distribution of within-mode action spread across keyframes.
    (b) Distribution of between-mode separation for multimodal keyframes.
    Distances measure the maximum displacement of fixed reference points on the gripper body between two goal poses, so both translation and orientation differences contribute in millimetres.
    }
    \label{fig:mode_geometry}
\end{figure}

\paragraph{Geometric structure of crowdsourced actions.}
To characterize the structure of the crowdsourced action distribution, we analyze the $K=10$ datasets at the keyframe level. For each keyframe, we compute pairwise distances between annotated actions using the maximum displacement of four fixed reference points on the gripper body under the two end-effector poses. This metric expresses both translational and rotational differences directly as physical displacement in millimetres. We cluster actions using complete-linkage agglomerative clustering with threshold $\tau_{\mathrm{EE}}=25.5$\,mm, calibrated to match the stringency of our corresponding joint-space analysis. Because complete linkage constrains the maximum within-cluster pairwise distance, each resulting cluster can be interpreted as a locally coherent action mode.

We quantify two complementary properties of the resulting action distribution (Fig.~\ref{fig:mode_geometry}). \emph{Within-mode spread} measures the median pairwise displacement among actions belonging to the same mode, aggregated to one value per keyframe. Its median values for Drawer, Insertion, Pouring, Sorting, and Switches are $17.0$, $2.8$, $9.4$, $7.9$, and $12.9$\,mm, respectively. \emph{Between-mode separation} measures the median displacement between mode medoids for multimodal keyframes. The corresponding medians are $43.9$\,mm for Drawer, $18.7$\,mm for Pouring, $122.8$\,mm for Sorting, and $201.5$\,mm for Switches; Insertion contains no multimodal keyframes under this criterion. These results show that crowdsourced supervision provides nontrivial geometric diversity even within individual modes, while the more strongly multimodal tasks additionally contain action modes separated by distances an order of magnitude larger than their local within-mode variation. Thus, increasing $K$ can provide both denser sampling within a valid mode and improved coverage of distinct valid modes.

\paragraph{Policy predictions at held-out states.} We further visualize how these differences in supervision appear in the learned policy by sampling 100 goal-pose predictions from policies trained with $K\in\{1,2,5,10\}$ at the same held-out keyframes (Fig.~\ref{fig:policy_preds}). Insertion is approximately unimodal, making it useful for isolating the within-mode densification effect. As shown in Fig.~\ref{fig:policy_preds}(a), low-width policies produce predictions with substantial bias and dispersion around the desired insertion pose. As $K$ increases, the predicted distribution becomes progressively more concentrated around the valid solution. Because there is no meaningful alternative task-level mode to cover at this keyframe, this behavior provides qualitative evidence that denser sampling within a mode is itself beneficial. Sorting, in contrast, contains keyframes with multiple valid next-action modes, such as choosing between the two remaining cubes. As shown in Fig.~\ref{fig:policy_preds}(b), predictions from low-width policies are comparatively diffuse and can smear through the region between the two valid modes. With increasing $K$, the modes become both more clearly separated and locally sharper; at $K=10$, the policy distinctly represents both alternatives rather than collapsing them into an intermediate prediction. Wider MAPS therefore provides both denser supervision within each solution mode and better coverage across solutions in this setting.

\paragraph{Relationship to task-level gains.}
The performance trends are consistent with these two effects compounding.
Averaged across the data regimes reported in Fig.~4, the improvement from
$K=1$ to $K=10$ is $+22$, $+32$, $+39$, and $+46$ percentage points for
Insertion, Pouring, Sorting, and Switches$^{-}$, respectively. The gains
therefore increase with the degree of multimodality in the task, broadly
matching the ordering of between-mode separation in the crowd labels.
Insertion provides evidence for the within-mode effect in relative
isolation, whereas on multimodal keyframes increasing $K$ simultaneously
increases within-mode density and the coverage of distinct action modes.
We therefore cannot disentangle the exact contribution of the two
mechanisms in the multimodal tasks, but the combined geometric and policy
prediction analyses are consistent with both contributing to the gains
from wider MAPS.

\subsection{Additional Results}\label{app:noise_baseline}
To isolate whether the performance gains from increasing dataset width $K$ stem from the geometric structure of crowdsourced actions, or can be replicated simply by adding appropriately noised versions of the executed action, which is cheaper and automated, we construct noise baseline datasets $\mathcal{D}_{\text{noise}}$.
Starting from a MAPS dataset $\mathcal{D}_{\text{MAPS}}$ with density $K=10$, we replace every unexecuted keyframe action $g \in G_t^{\neg\text{exec}}$ with a synthetic sample drawn from a Gaussian centered at the corresponding executed action:
\begin{equation} \label{eq:noise_baseline}
\tilde{g}_t^{(k)} = g_t^{\text{exec}} + \epsilon, \quad \epsilon \sim \mathcal{N}(\mathbf{0},\, \hat{\Sigma}), \quad \forall\, g_t^{(k)} \in G_t^{\neg\text{exec}},\ t \in \mathcal{K},
\end{equation}
where $\hat{\Sigma} = \operatorname{diag}(\hat{\sigma}_1^2, \dots, \hat{\sigma}_7^2)$ is a diagonal covariance estimated from the original crowd labels. Concretely, we compute $\hat{\Sigma}$ by (i)~extracting the 7-DoF joint-state components of all keyframe actions (both executed and unexecuted), (ii)~clustering them per keyframe via MeanShift to identify distinct action modes, and (iii)~computing the within-cluster sample variance across all clusters:
\begin{equation} \label{eq:noise_variance}
\hat{\sigma}_j^2 = \frac{1}{|\mathcal{C}|} \sum_{c \in \mathcal{C}} \operatorname{Var}\!\left[\left\{ g^{(k)}_j \right\}_{k \in c}\right], \quad j = 1, \dots, 7,
\end{equation}
where $\mathcal{C}$ is the set of all MeanShift clusters containing more than one action, and $g^{(k)}_j$ denotes the $j$-th pose component of the $k$-th action in cluster $c$. The resulting noise baseline $\mathcal{D}_{\text{noise}}$ preserves the dataset structure (same observations, same number of actions per keyframe, same executed actions) but replaces the human-provided geometric diversity with diagonal-covariance Gaussian perturbations. 

We leverage information from the crowdsourced action labels to construct $\mathcal{D}_{\text{noise}}$ because (1) it is difficult to decide arbitrarily the variance of the perturbations, such that it forms a fair baseline, and (2) since we are using statistics provided by the crowdsourcing process, this formulation gives us a stronger baseline than adding arbitrary noise, especially for tasks with only one keyframe variation (Insertion and Switches).

As shown in Fig.~\ref{fig:keypose_results}, the Gaussian-noise baseline sometimes outperforms single-action supervision ($K=1$), but consistently underperforms genuine crowdsourced MAPS at $K=10$ on the reported task averages. Moreover, its performance exhibits negative scaling in several settings: as training episodes increase, success drops from $88\%$ to $52\%$ on Insertion and from $60\%$ to $24\%$ on Sorting. These results indicate that matching action density through Gaussian perturbations, even when calibrated using crowd annotations, cannot reproduce the benefits of genuine crowdsourced alternatives.

\end{document}